\documentclass[11pt,journal,draftcls,onecolumn]{IEEEtran}
\ifCLASSOPTIONcompsoc
\else
\fi
\ifCLASSINFOpdf
\else
\fi

\usepackage{graphicx,amssymb,amsmath,algpseudocode,algorithm,epstopdf,subfig}
\usepackage{slashbox}
\usepackage{array}
\usepackage{gensymb}
\usepackage{multirow}
\usepackage{url}
\usepackage{xcolor}

\newcolumntype{L}[1]{>{\raggedright\let\newline\\\arraybackslash\hspace{0pt}}m{#1}}
\newcolumntype{C}[1]{>{\centering\let\newline\\\arraybackslash\hspace{0pt}}m{#1}}
\newcolumntype{R}[1]{>{\raggedleft\let\newline\\\arraybackslash\hspace{0pt}}m{#1}}

\begin{document}

%
\title{Nasal Patches and Curves for Expression-robust {3D} Face Recognition}
%
%
%
%

\author{Mehryar~Emambakhsh~and~Adrian~Evans
\IEEEcompsocitemizethanks{\IEEEcompsocthanksitem M. Emambakhsh was with the Department of Electronic and Electrical Engineering, University of Bath, Bath, UK. E-mail: mehryar{\_}emam@yahoo.com\protect
}
\IEEEcompsocitemizethanks{\IEEEcompsocthanksitem A.N. Evans is with the Department of Electronic and Electrical Engineering, University of Bath, Bath,
UK. E-mail: A.N.Evans@bath.ac.uk\protect}
\thanks{}}

\IEEEcompsoctitleabstractindextext{%
\begin{abstract}
The potential of the nasal region for expression robust {3D} face recognition is thoroughly investigated by a novel five-step algorithm. First, the nose tip location is coarsely detected and the face is segmented, aligned and the nasal region cropped. Then, a very accurate and consistent nasal landmarking algorithm detects seven keypoints on the nasal region. In the third step, a feature extraction algorithm based on the surface normals of Gabor-wavelet filtered depth maps is utilised and, then, a set of spherical patches and curves are localised over the nasal region to provide the feature descriptors. The last step applies a genetic algorithm-based feature selector to detect the most stable patches and curves over different facial expressions. The algorithm provides the highest reported nasal region-based recognition ranks on the FRGC, Bosphorus and BU-3DFE datasets. The results are comparable with, and in many cases better than, many state-of-the-art {3D} face recognition algorithms, which use the whole facial domain. The proposed method does not rely on sophisticated alignment or denoising steps, is very robust when only one sample per subject is used in the gallery, and does not require a training step for the landmarking algorithm.\\
{\color{blue}{\url{https://github.com/mehryaragha/NoseBiometrics}}}
\end{abstract}


}

\maketitle
\bibliographystyle{ieeetr}

\IEEEdisplaynotcompsoctitleabstractindextext

%
\IEEEpeerreviewmaketitle

\section{Introduction}
%
%

%
%
%
%
\IEEEPARstart{W}{hile} much previous research on expression invariant {3D} face recognition has focused on modelling expressions and detecting expression insensitive facial parts, there have been relatively few studies evaluating the potential of the nasal region for addressing this issue. Despite this, the nose has a number of salient features that make it suitable for expression robust recognition. It can be easily detected, due to its discriminant curvature and convexity \cite{Chang:2006}, is difficult to hide without attracting suspicion \cite{Emambakhsh:2011,Moorhouse:2009}, is relatively stable over various facial expressions (\cite{Chang:2006,Emambakhsh:2013,Wang:20082,Alyuz:2010,Drira:2009,Dibeklioglu:2009,Mian:2007}) and is rarely affected by unintentional occlusions caused by hair and scarves. Although it has been reported that the {2D} image of the nose has too few discriminant features to be used as a reliable region for human identification \cite{Zhao:2003}, its {3D} surface has much undiscovered potential. This paper further investigates the {3D} nasal region for human identity authentication and verification purposes and presents a novel algorithm that provides very high discriminant strength, comparable with recent {3D} face recognition algorithms, which use the whole facial domain.

The proposed approach is based on a very consistent and accurate landmarking algorithm, which overcomes the issue of robust segmentation of the nasal region. The algorithm first finds an approximate location of the nose tip and then finely tunes its location, while accurately determining the position of the nasal root and detecting the symmetry plane of the face. Next, the locations of three sets of landmarks are found: subnasale, eye corners and nasal alar groove. These landmarks are utilised on feature maps created by applying multi-resolution Gabor wavelets to the surface normals of the depth map. Two types of feature descriptors are used: spherical patches and nasal curves. Feature selection is then performed using a heuristic genetic algorithm (GA) and, finally, the expression-robust feature descriptors are applied to the well-known and widely used {3D} Face Recognition Grand Challenge (FRGC) \cite{Phillips:2005}, Bosphorus \cite{Savran:2008} and Binghamton University 3D Facial Expression (BU-3DFE) \cite{Yin:2007} datasets.

Results show the algorithm's high potential to recognise nasal regions, and hence faces, over different expressions, with very few gallery samples per subject. The highest rank-one recognition rates (R$_1$RR) achieved are: 1) a R$_1$RR of 97.9\% and equal error rate (EER) of 2.4\% for FRGC v2.0 and receiver operator characteristic (ROC) III experiments, respectively; 2) a R$_1$RR of 98.45\% and 98.5\% for FRGC's neutral vs. neutral and neutral vs. non-neutral samples, respectively; 3) a R$_1$RR of 96.2\% when one gallery sample per subject is used for the FRGC dataset (482 gallery samples (subjects) vs. 4330 probe samples); 4) a R$_1$RR of 95.35\% for the Bosphorus dataset when 2797 scans of 105 subjects are used as probes and the set of 105 neutral scans (one per subject) is used as the galley.

The remainder of the paper is organized as follows. After the literature review provided in section~\ref{sec:lit_review}, the alignment and nasal region cropping steps, followed by the nasal region landmarking, are detailed in section \ref{sec:nosetipdetecinitlandmarkingnasal}. The feature extraction algorithm is described in section \ref{sec:featextract} and section \ref{sec:featdescription} explains the feature descriptors used. The feature selection algorithm is detailed in section \ref{sec:featselection} and experimental results, including a thorough comparison with previous work, is provided in section \ref{sec:expresults}. Finally, conclusions are given in section \ref{sec:conc}.

\subsection{Scientific contribution and comparison with previous work}
The major contribution of this paper is a novel surface normal-based recognition algorithm that provides a thorough evaluation of the recognition potential of the {3D} nasal region. The results achieved are not only better than previous {3D} nose recognition algorithms but also higher than many recognition algorithms that employ the whole face. The algorithm employs a novel, training-free, highly consistent and accurate landmarking algorithm for the nasal region and a robust feature space, based on the response of Gabor wavelets to surface normal vectors, is also introduced. To localise the expression robust regions on the nose a heuristic GA feature selection is applied to two different geometrical feature descriptors. Because of the smoothing effects of the Gabor wavelets, there is no need for sophisticated denoising algorithms. Indeed, only simple median filtering is required for the surface normals, even with noisy datasets such as the FRGC Spring 2003 folder. An additional advantage of the proposed approach is that a fast Principal Component Analysis (PCA)-based self-dependent method can be employed for facial pose correction. This eliminates the need for sophisticated pose correction algorithms or reference faces for fine tuning the alignment.

The proposed approach significantly extends our previous work~\cite{Emambakhsh:2013} in which the nasal landmarking and recognition was performed on the depth map. This paper increases the number of landmarks and their detection accuracy and presents new feature extraction and selection algorithms.
The work is inspired by recent algorithms on utilising facial normal vectors in {3D} \cite{Mohammadzade:2013} and regional normal vectors \cite{Li:2014}. To compare the new algorithms with previous approaches which used similar methodologies, the application of normals, computed over the nasal surface, is used for identification as well as the verification scenario. By using multi-resolution Gabor wavelets the ability of the algorithm to handle more noisy samples is enhanced, providing higher R$_1$RR than the approach of Li \emph{et al.} \cite{Li:2014}, which excluded the noisy FRGC Spring 2003 samples. This work also extends the application of facial curves, introduced as feature descriptors by Berretti \emph{et al.} (\cite{Berretti:2010} and \cite{Berretti:2013}), to nasal spherical patches, producing a R$_1$RR increase of $>2\%$, and showing a higher class separability for the spherical patches than for curves for {3D} face recognition.

\section{Recent literature review \label{sec:lit_review}}
Robustness against the deformations caused by facial expressions has been a popular research topic in {3D} face recognition. The face is a non-rigid object and therefore {3D} matching techniques for rigid objects, such as the iterative closest point (ICP) algorithm~\cite{Besl:1992}, can become trapped in local minima and fail to provide accurate matching scores.

An empirical approach to deal with the variations caused by expressions is to capture a range of facial expressions for each subject and store them in the gallery \cite{Bowyer:2006}. Then, the facial biometric features of each test subject can be compared with all the stored expressions and a decision made on the identity of the subject. This method has numerous disadvantages: capturing a range of facial expressions for each subject is not always straightforward and requires a high storage capacity per subject. In addition, facial expressions will not necessarily remain constant and may differ between the test and gallery captures \cite{Bowyer:2006}.

One approach to overcome this problem is to use computer graphics algorithms to artificially create different expressions for each facial capture. In \cite{Osaimi:2009}, expressions are learned using PCA eigenvectors and then used to re-generate the expressions on the probe samples. Although this approach does not require multiple samples per subject in the gallery, it is still vulnerable to the number of training samples used to model the facial expressions. Also, a universal definition of facial expression for all subjects still remains to be found \cite{Bowyer:2006} and the need to classify the expression types prior to face recognition increases the computational complexity.

Another approach is to employ region-based methods, in which the least variant parts of the face over different expressions are detected using facial segmentation \cite{Chang:2006, Spreeuwers:2011, Alyuz:2010, Queirolo:2010} or extracted using their expression invariant capabilities \cite{Mohammadzade:2013, Wang:2010, Wang:20082, Li:2014}. Spreeuwers proposes a multiple regional approach based on a PCA-Linear Discriminant Analysis (LDA) feature extraction method~\cite{Spreeuwers:2011}. In regional recognition, scores from a combination of different masks on the nose, cheek, forehead, chin and mouth are fused to finalise the decision making.

Aly{\"u}z \emph{et al.} use a regional registration algorithm in conjunction with LDA classifiers, giving an expression robust {3D} face recognition approach~\cite{Alyuz:2010}. They also demonstrate that the nasal region has a high discriminatory power. A focus on integrating multiple regions is provided by Queirolo \emph{et al.}~\cite{Queirolo:2010} in which four regions (the upper face image, the whole face and two nasal regions) are segmented and stored for the gallery sessions before matching is performed using a novel matching criterion, called the surface interpenetration measure, and simulated annealing.

Using facial curves is another popular approach to {3D} face recognition that can be categorised as a subset of regional algorithms. Drira \emph{et al.} use the intersections of planes with the facial surface to define a set of radial curves which pass through the nose tip, and then perform a quality assessment in order to handle missing data and occlusions~\cite{Drira:2013}. Another curve-based algorithm is proposed by Berretti \emph{et al.}~\cite{Berretti:2013}. First, keypoints are detected on the facial surface and then the least variant curves on the face are selected using a statistical model and matched with those in the gallery. As an extension to curves, isogeodesic stripes centralised on the nose tip are used in an expression invariant {3D} face recognition method that employs a novel descriptor, termed the {3D} weighted walkthroughs, to quantify the differences between corresponding stripes \cite{Berretti:2010}.
In another curve-based approach, Drira \emph{et al.} find geodesic curves on the nasal region for a subset of the FRGC dataset~\cite{Drira:2009}.

To overcome the sensitivity of holistic face recognition algorithms to expression variations, Mian \emph{et al.} propose a landmark-based method, in conjunction with a localised feature descriptor that incorporates the {2D} texture and {3D} point clouds. In an alternative approach, Wang \emph{et al.} apply shape difference boosting to the Bosphorus dataset to learn the expressions and identify those facial regions which remain constant over different expressions~\cite{Wang:2010}. Instead of using depth or the point coordinates for {3D} registration, Mohammadzade~\emph{et al.} use the surface normals of the points in conjunction with a Fisher's discriminant paradigm \cite{Mohammadzade:2013}. This approach selects the normals which maximise the concentration of within-class scatter while simultaneously maximising the between-class distribution. Recently, Li \emph{et al.} proposed local normals histograms, captured from multiple rectangular regions on the face, to set up an expression-robust feature space and use a novel sparse classifier to perform the matching~\cite{Li:2014}.

Despite the robustness of these algorithms against facial expressions, they often rely on accurate and consistent facial segmentation, which is not a straightforward task in {3D}. To address this issue, some researches have focused on the nasal region, which shows high consistency over different expressions. For example, in one of the first investigations on {3D} nose recognition, Chang~{\emph{et al.}} initially segment the face into different non-overlapping regions, using the curvature information~\cite{Chang:2006}. Then, three overlapping nasal regions are detected and stored in the gallery. The same regions are segmented in the probe images and matched using the ICP algorithm. Wang {\emph {et al.}} propose the use of local shape difference boosting for {3D} face recognition and also apply the boosting algorithm to different nasal regions~\cite{Wang:20082}. The regions are cropped using the intersection of spheres of radius $r$, centred on the nose tip, with the face surface. When the value of $r$ was increased, the recognition ranks reached a maximum and then plateaued. A combination of the nasal region, forehead and eyes are used for a {2D}/{3D} face recognition by Mian {\emph {et al.}}~\cite{Mian:2007}. A modified ICP algorithm is used for matching, in conjunction with a pattern rejector based on spherical face representation (SFR) and shift-invariant feature transform (SIFT) \cite{Lowe:2004}, producing high recognition ranks on the FRGC dataset, in particular for the neutral probes. Dibeklio\u{g}lu \emph{et al.} used the Dijkstra algorithm to segment the nose and evaluated the performance using a subset of the Bosphorus dataset \cite{Dibeklioglu:2009}.

\begin{figure}[!tb]
\centering
\subfloat[]{\includegraphics[width=0.5\textwidth]{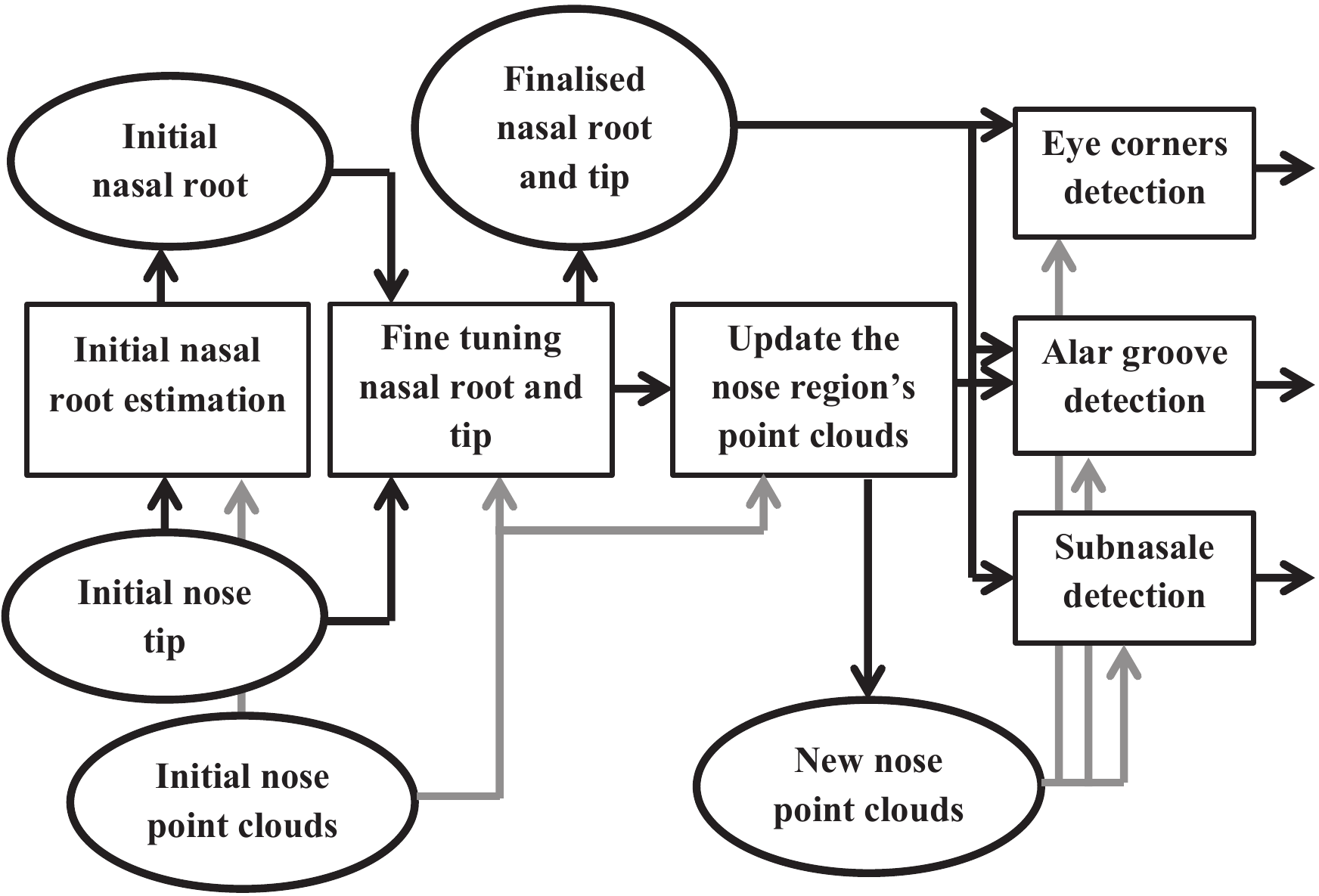}}\\
\subfloat[]{\includegraphics[width=0.3\textwidth]{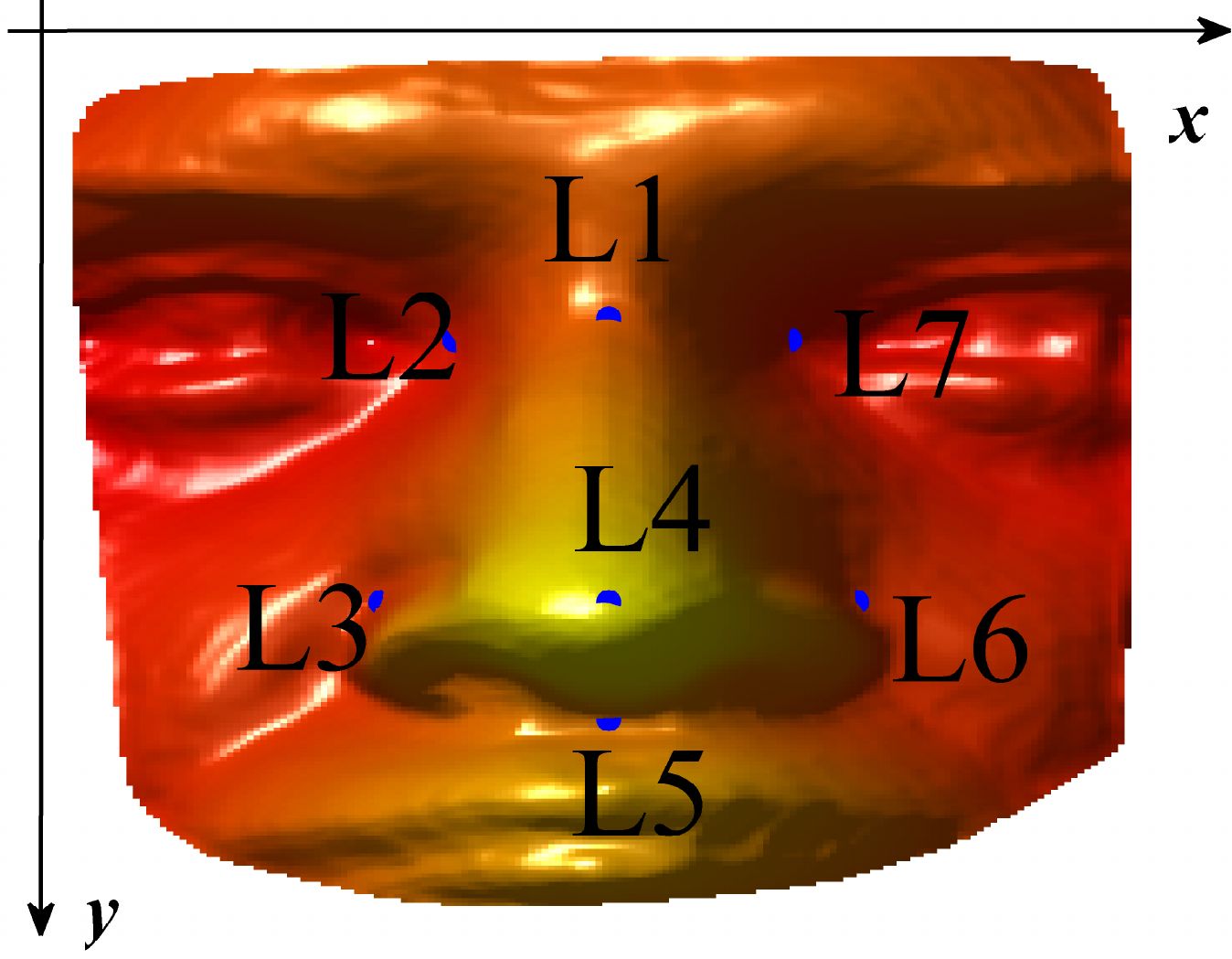}}
\caption{(a) The landmarking algorithm steps in a block diagram; (b) The naming convention for the nasal landmarks in our work.}
\label{Fig:LandBlockLandmarksNames}
\end{figure}

\section{Preprocessing and nasal region landmarking}
\label{sec:nosetipdetecinitlandmarkingnasal}
The algorithm explained in \cite{Emambakhsh:2013} is used to crop the face. Next, median filtering with a $2.5\times2.5$~mm$^2$ mask size is applied twice on the cropped face. The image is then resampled to a uniform grid with $0.5$~mm/pixel horizontal and vertical resolutions using Delaunay triangulation and aligned using the iterative PCA algorithm \cite{Mian:2007}. The aligned face is then intersected with three cylinders to crop the nasal region, according to \cite{Emambakhsh:2013}. The depth map of the cropped nasal region is again median filtered with a $2.5\times2.5$~mm$^2$ mask to further smooth its surface and decrease the spike noise effects. The block diagram in Fig. \ref{Fig:LandBlockLandmarksNames}-a shows how the landmarks in Fig. \ref{Fig:LandBlockLandmarksNames}-b are detected.

\subsection{Local minima detector, nose tip re-localisation, nasal root and subnasale detection}
\begin{figure}[!tb]
\centering
\subfloat[]{\includegraphics[width=0.25\textwidth]{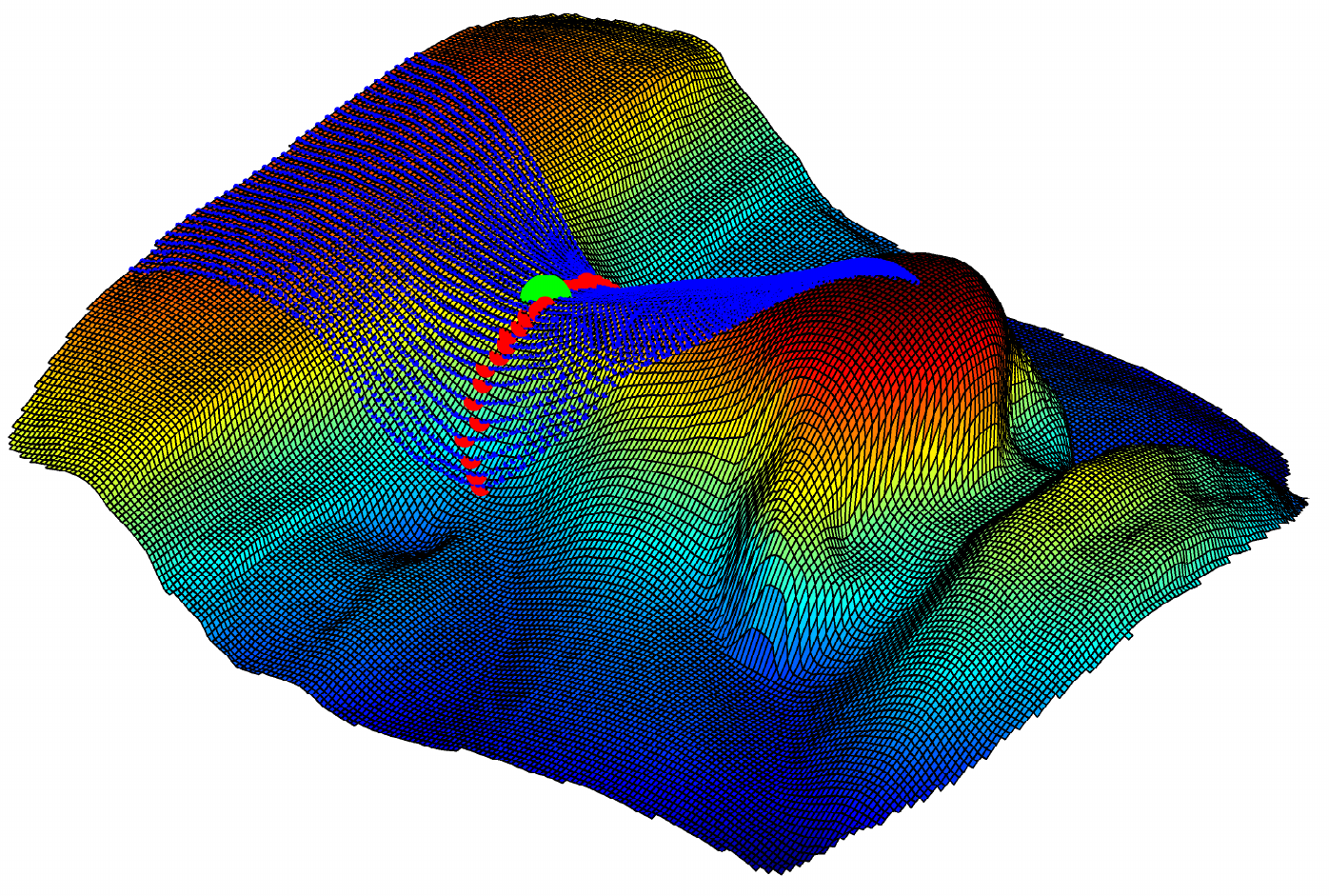}}
\subfloat[]{\includegraphics[width=0.25\textwidth]{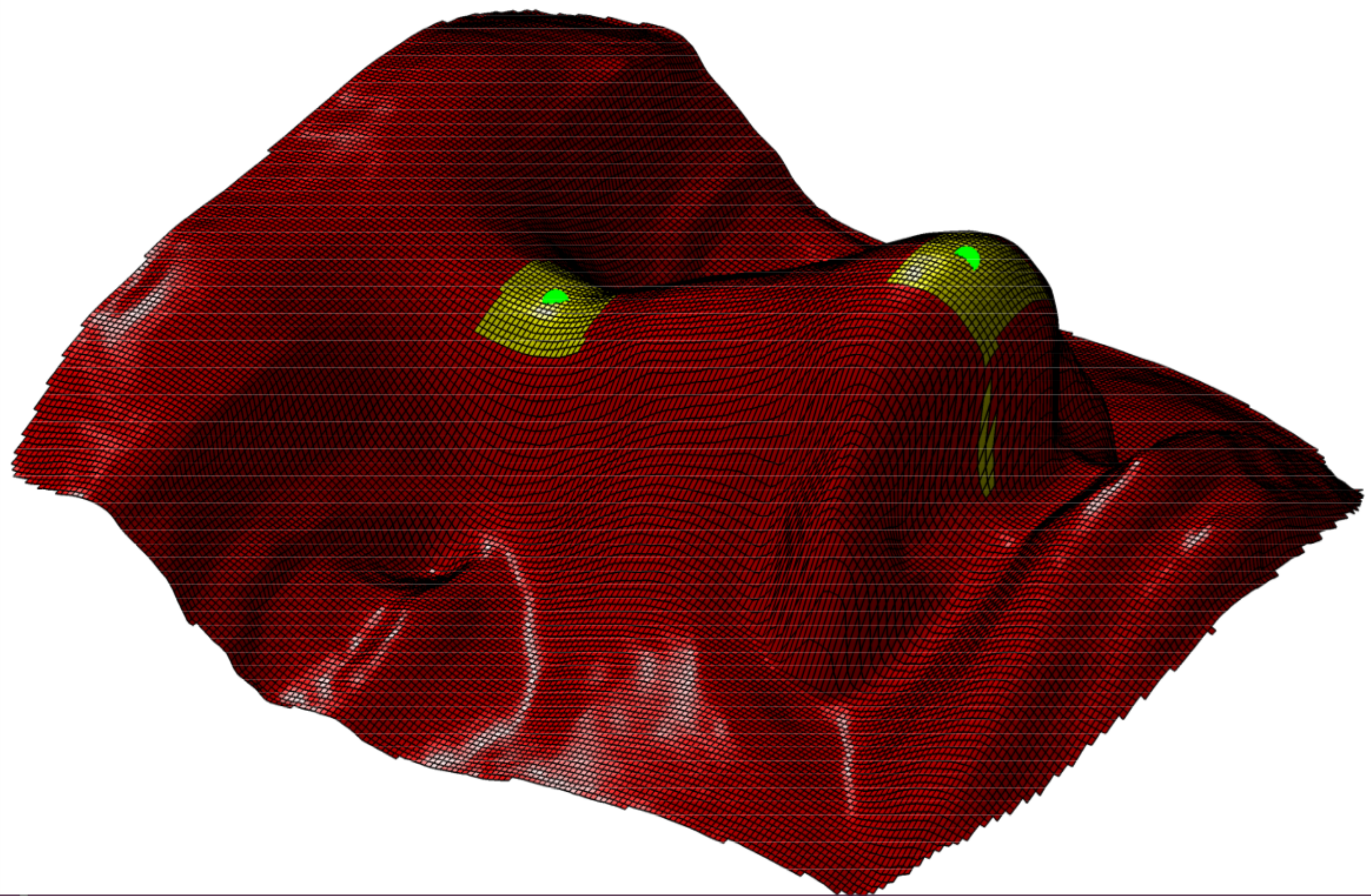}}\\
\subfloat[]{\includegraphics[width=0.25\textwidth]{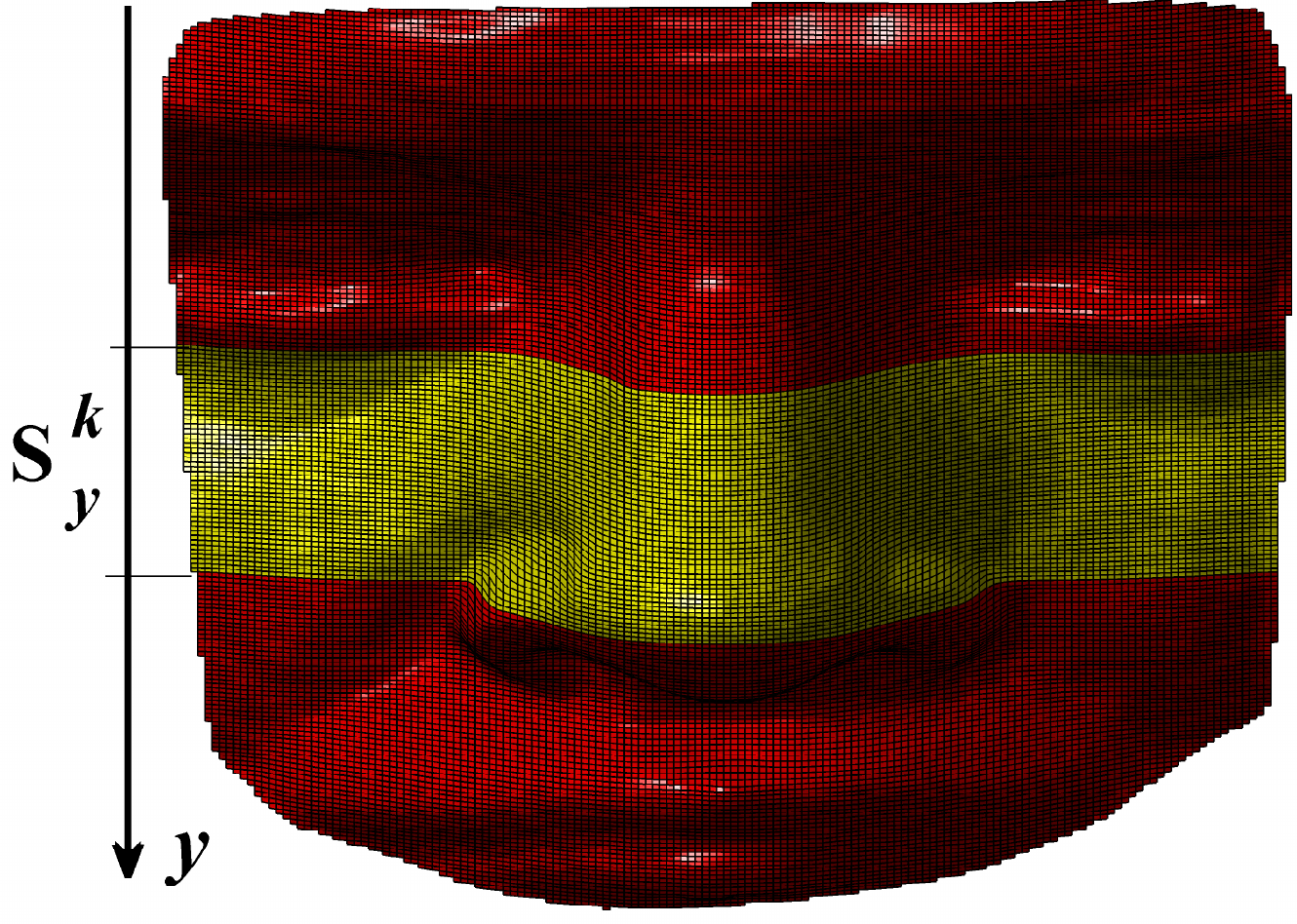}}
\caption{(a) Localisation procedure for the nasal root, $\bf{L1^0}$, shown in green: the blue curves and red points represent planes intersections and their minima, respectively. (b) The $5\times5$ mm$^2$ RoIs around ${\bf{L1}}^i$ and ${\bf{L4}}^j$. (c) The horizontal strip ${\bf S}^k_y$ used in (\ref{eq:MaxSymCol}).}
\label{Fig:SaddleDetecROITIPSADDLEHorizStrip}
\end{figure}

First, an initial position of the nasal root ($\bf{L1^0}$) is detected by \cite{Emambakhsh:2013}. Then, the location of the nose tip ($\bf{L4^0}$), found in section \ref{sec:nosetipdetecinitlandmarkingnasal}, is more finely tuned. Various planes, passing through $\bf{L4^0}$ with normals $\cos (\theta_i)\hat{a}_x + \sin (\theta_i)\hat{a}_y$ are intersected with the nose surface, where $\theta_i$ is the angle of the $i^{th}$ plane with the $y$-axis, and $\hat{a}_x = [1, 0]$ and $\hat{a}_y = [0, 1]$ are the unit vectors along the $x$ and $y$ axes, respectively. This process results in several curves on the nasal region, shown in Fig. \ref{Fig:SaddleDetecROITIPSADDLEHorizStrip}-a.

\begin{figure}[!tb]
\centering
\includegraphics[width=0.5\textwidth]{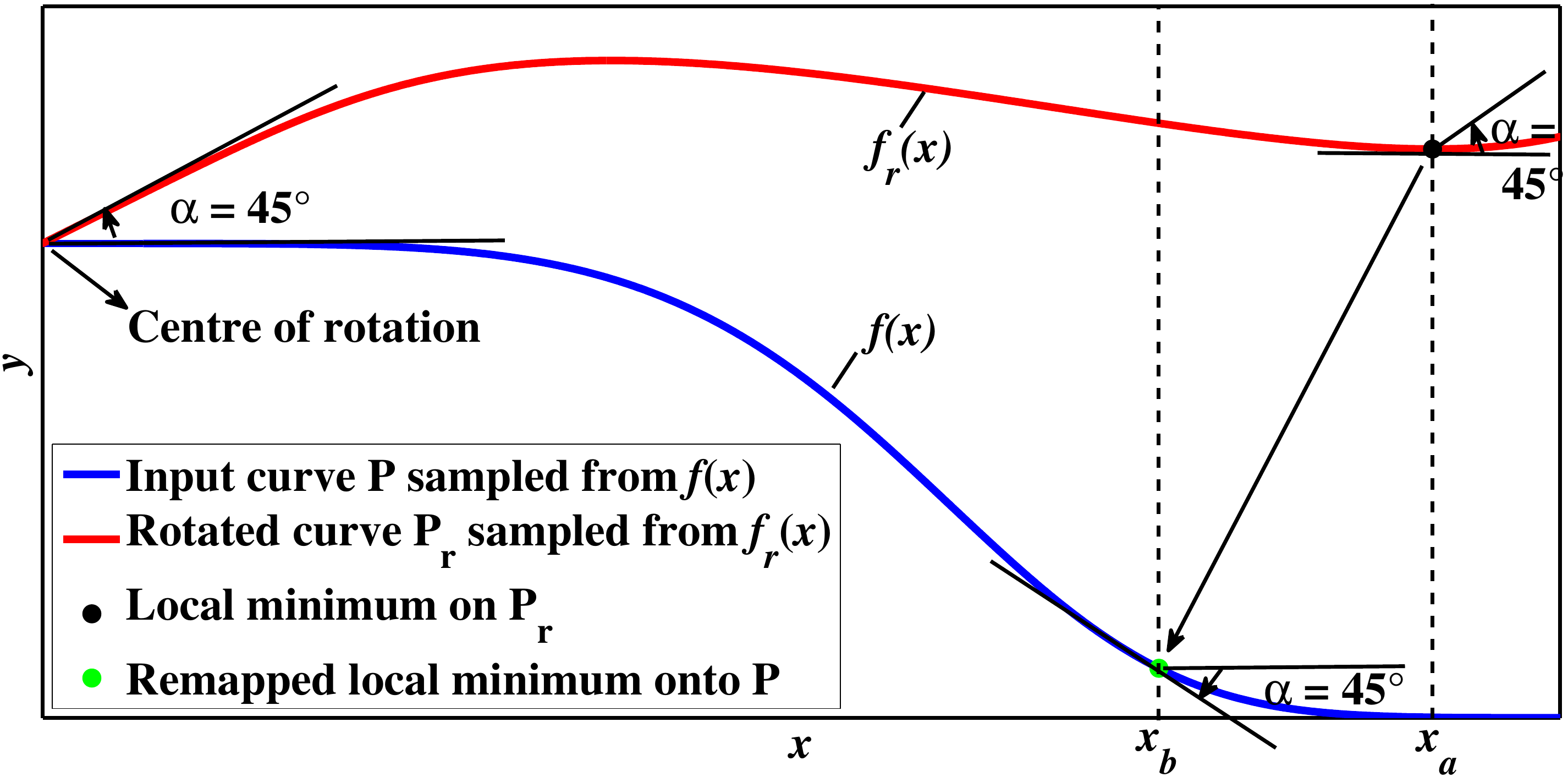}
\caption{The blue curve is strictly decreasing without any minima, while its $45\degree$ rotation (red curve) has a distinctive minimum. The interpretation of this procedure in the continuous space is $\alpha=\arctan\left(\left|\frac{df(x)}{dx}|_{x = x_b}\right|\right)$ while $\frac{df_r(x)}{dx}|_{x = x_a}=0$.}
\label{Fig:CurveMin}
\end{figure}

The proposed landmarking algorithm relies on a minima detector, which finds a set of minima on rotated versions of the curves and then maps them to the original curve. The rotation is required because some of the original curves are strictly decreasing functions that do not have an actual minimum. Assuming ${\bf{P}} = [{\bf{X_f}}, {\bf{Y_f}}]$ is a $K\times2$ matrix representing points of a curve, instead of directly differentiating ${\bf{P}}$ to find the minima on the curves, as proposed by Segundo \emph{et al.} \cite{Segundo:2010}, the curves are first rotated in the $z$-axis (roll direction) by an angle $\alpha$ around a given point on the curve. This operation is given by,
\begin{equation}
	\left\{
	\begin{array}{l}
	{\bf{P}_r} = {\bf{P}} \times \left[ \begin{array}{cc}
\cos \alpha  & \sin \alpha \\
-\sin \alpha & \cos \alpha \end{array} \right] = [{\bf{X_f^\alpha}}, {\bf{Y_f^\alpha}}]  \\
{\bf{MIN}} =  V_{n, \alpha}({\bf{P}_r})
\end{array}\right.,
\label{eq:mindetector}
\end{equation}
\noindent where ${\bf{P}_r}$ is the rotated version of ${\bf{P}}$ and the function $V_{n, \alpha}(.)$ finds the location of the $n$ smallest local minima on ${\bf{P}_r}$ and then remaps them to the original curve $\bf P$ using the rotation angle $\alpha$. The output ${\bf{MIN}}$ is an $n \times 2$ matrix, containing the locations of the $n$ local minima. $V_{n, \alpha}(.)$ computes the first order differentiation (first order difference in discrete space) of $\bf P_r$, which is then given to the {\em signum} function to detect its sign changes. This finds the locations of all the local minima in ${\bf{P}_r}$, which are then sorted based on their value in ascending order and the $n$ lowest are selected and rotated back to the original curve using $\alpha$. Fig.~\ref{Fig:CurveMin} shows an example of this procedure for $n=1$ (the global extremum).

The value allocated to $\alpha$ should be small enough in order to preserve the single-valued functionality of ${\bf{X_f^\alpha}} \rightarrow {\bf{Y_f^\alpha}}$, i.e. each projection of any point on ${\bf{P}_r}$ to the horizontal axis should correspond to only one point on the vertical axis. Based on the type of landmark to be extracted, the value of $\alpha$ is chosen using trial and error.
The minima detector operator of (\ref{eq:mindetector}) is applied to each curve in Fig. \ref{Fig:SaddleDetecROITIPSADDLEHorizStrip}-a as follows,

\begin{equation}
{{\bf{SMIN}}_m} =  V_{1, \beta}({{\bf{S}}^m_{\gamma_m}})
\label{eq:forsaddlecand}
\end{equation}

\noindent in which, ${{\bf{S}}^m_{\gamma_m}}$ represents the $m^{th}$ curve, which is rotated by $\gamma_m$ around $\bf{L4^0}$.
${{\bf{SMIN}}_m}$ represents the location of the global minimum for each curve, which itself, constitutes a curve whose global maximum gives $\bf{L1^0}$, the initial location of the nasal root. Figure~\ref{Fig:SaddleDetecROITIPSADDLEHorizStrip} shows the set of curves ${{\bf{S}}^m_\gamma}$ (in blue), their minima (in red) and the maximum of the minima (green).

The nasal root and tip locations ($\bf{L1^0}$ and $\bf{L4^0}$) may be slightly inaccurate due to the depth variations caused by the noise and facial expressions. In order to improve the accuracy of their locations, for the points situated on a 5 $\times$ 5 mm$^2$ area (shown in Fig. \ref{Fig:SaddleDetecROITIPSADDLEHorizStrip}-b) around the nasal root and saddle, the following angular deviation is calculated,

\begin{equation}
	\theta_z^k = \arccos{\frac{\left|[L4^j_x - L1^i_x, L4^j_y - L1^i_y] \cdot \hat{a}_y\right|}{\left|[L4^j_x - L1^i_x, L4^j_y - L1^i_y]\right|}},
\label{eq:}
\end{equation}

\noindent in which, $[L1^i_x, L1^i_y]$ and $[L4^j_x, L4^j_y]$ are the $[x, y]$ projections of the two pairs of points ${\bf{L1}}^i$ and ${\bf{L4}}^j$ ($i = 1, \ldots, I$, $j = 1, \ldots, J$ and $k=1, \ldots, J \times I$), which are selected from the overall $I$ and $J$ points on the region of interest (RoI) from the nasal root and tip regions, respectively, see Fig. \ref{Fig:SaddleDetecROITIPSADDLEHorizStrip}-b. $\theta_z^k$ is used to rotate the nose region in the roll direction and around ${\bf{L4}}^j$. Then the image is divided into the left and right halves. Assuming the rotated nasal region is translated so that the nose tip ${\bf{L4}}^j$ is at the origin, for the $y$-axis indices within the strip ${\bf{S}}^k_y$ shown in Fig. \ref{Fig:SaddleDetecROITIPSADDLEHorizStrip}-c (computed using $\theta_z^k$), the objective function $E^k$ is calculated by


\begin{equation}
E^k({\bf{L1}}^i, {\bf{L4}}^j) =
\operatorname*{\max}_{y \in {\bf{S}}^k_y} \left( \sum_{x}\left|{\bf{Z}}^k_{L} - {\bf{Z}}^k_{R} \right| \right)
\label{eq:MaxSymCol}
\end{equation}

\noindent in which ${\bf{Z}}^k_{L}$ and ${\bf{Z}}^k_{R}$ are the depth maps of the flipped and cropped left and right sides of ${\bf{L4}}^{j}$ and ${\bf{L1}}^{i}$, respectively.
The two points ${\bf{L1}}^{opt}$ and ${\bf{L4}}^{opt}$ that minimise $E^k$ have the most similar values of ${\bf{Z}}^{opt}_L$ and ${\bf{Z}}^{opt}_R$ and their projections onto the $x$ axis (${L1^{opt}_x}$ and ${L4^{opt}_x}$) correspond to the $x$ values of the accurate nose tip and root locations such that,

\begin{equation}
\theta_z^{opt} = \operatorname*{\arg min}_{\theta_z^k} E^k({\bf{L1}}^i, {\bf{L4}}^j).
\label{eq:newOpt}
\end{equation}

\begin{figure}[!tb]
\centering
\subfloat[]{\includegraphics[width=0.2\textwidth]{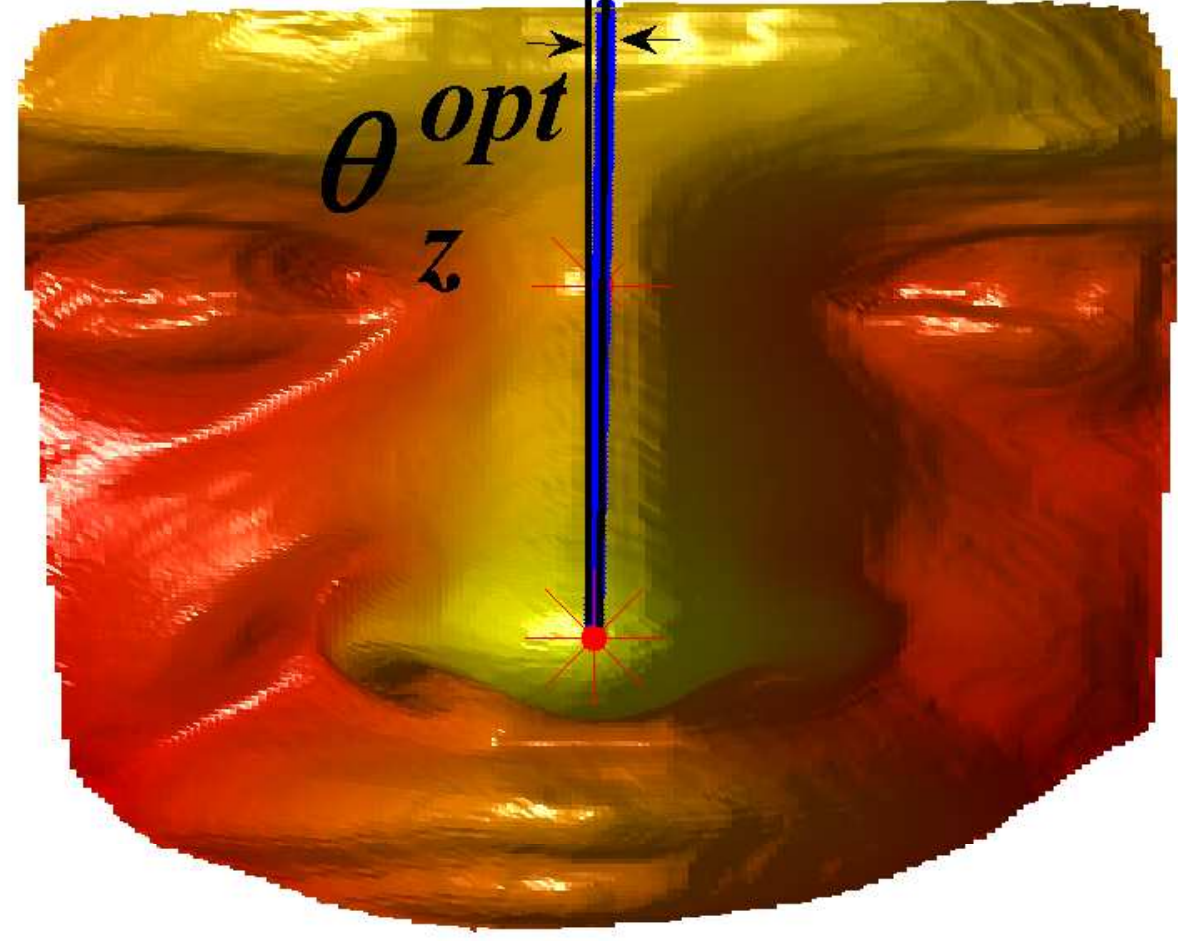}}
\subfloat[]{\includegraphics[width=0.2\textwidth]{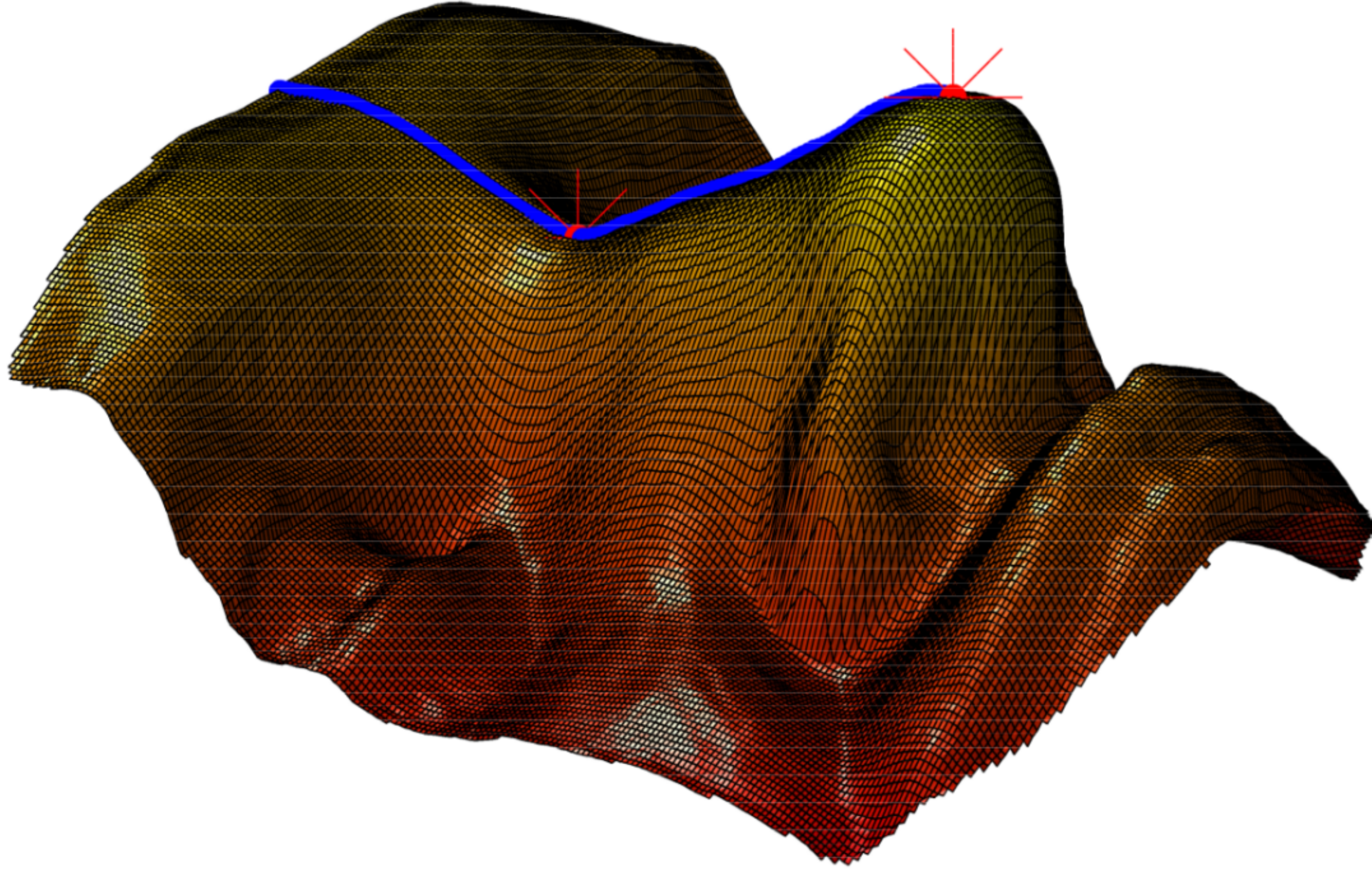}}\\
\subfloat[]{\includegraphics[width=0.20\textwidth]{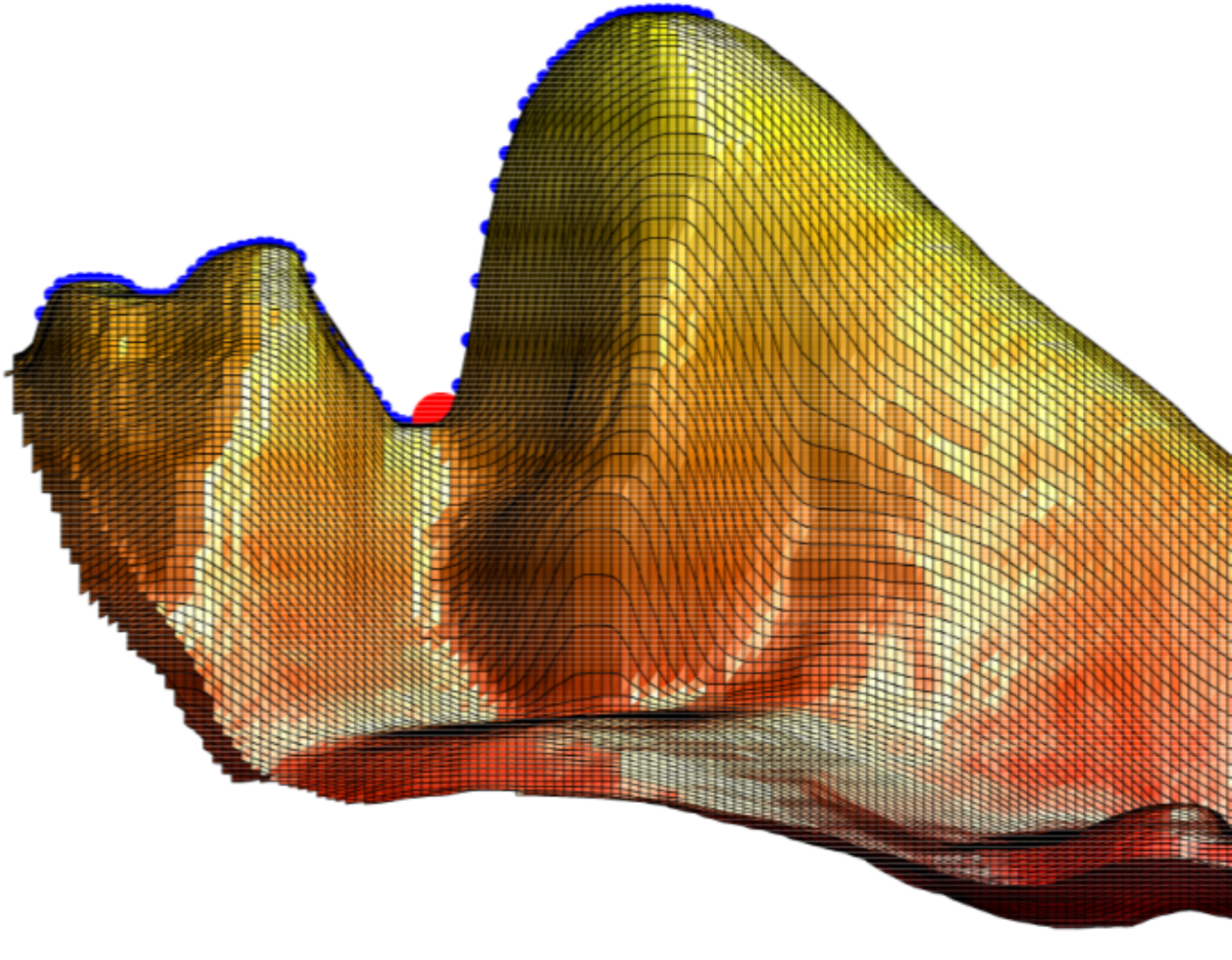}}
\caption{Nasal root, tip and subnasale detection: (a) Updating the nasal region using $\theta_z^{opt}$. (b) The maximum and minimum of a curve connecting the optimum  $[{L4^{opt}_x}, {L4^{opt}_y}]$ and $[{L1^{opt}_x}, {L1^{opt}_y}]$ (blue curve) are used as ${\bf{L4}}$ and ${\bf{L1}}$, respectively (red points); (c) Blue points: symmetry plane intersection; Red point: the lowest minimum, detected as subnasale.}
\label{Fig:TipSadNearlythere}
\end{figure}

This is an example of a "min-max" optimisation, which finds the best worst case for the optimum \cite{Aissi:2009}.
A plane passing through ${\bf{L1}}^{opt}$ and ${\bf{L4}}^{opt}$ is then intersected with the nose surface, with normal vector $[\cos \theta_z^{opt}, \sin \theta_z^{opt} ]$, see Fig.~\ref{Fig:TipSadNearlythere}-b. The locations of the maximum and minimum of the resulting curve are the positions of ${L1^{opt}_y}$ and ${L4^{opt}_y}$. This procedure is illustrated in Fig.~\ref{Fig:TipSadNearlythere}.
${\bf{L4}}=[{L4^{opt}_x}, {L4^{opt}_y}, {L4^{opt}_z}]$ and ${\bf{L1}}=[{L1^{opt}_x}, {L1^{opt}_y}, {L1^{opt}_z}]$ give the final locations of the nose tip and nasal root, respectively.

The points on the same curve, which are located below the nasal tip $\bf{L4}$ (shown in Fig.~\ref{Fig:TipSadNearlythere}-c)
are then rotated around $\bf{L4}$ by an angle $\phi$. The location of the lowest minimum of the resulting curve ($\bf{S}_{\phi}$) provides the subnasale $\bf{L5}$ after applying (\ref{eq:mindetector}), i.e. ${\bf{L5}} = V_{1, \phi}({\bf{S}}_{\phi})$. Finally, $\theta_z^{opt}$ is used to update the nose region and correct the pose by applying a roll directional rotation around ${\bf{L4}}$.

\begin{figure}[!tb]
\centering
\subfloat[]{\includegraphics[width=0.2\textwidth]{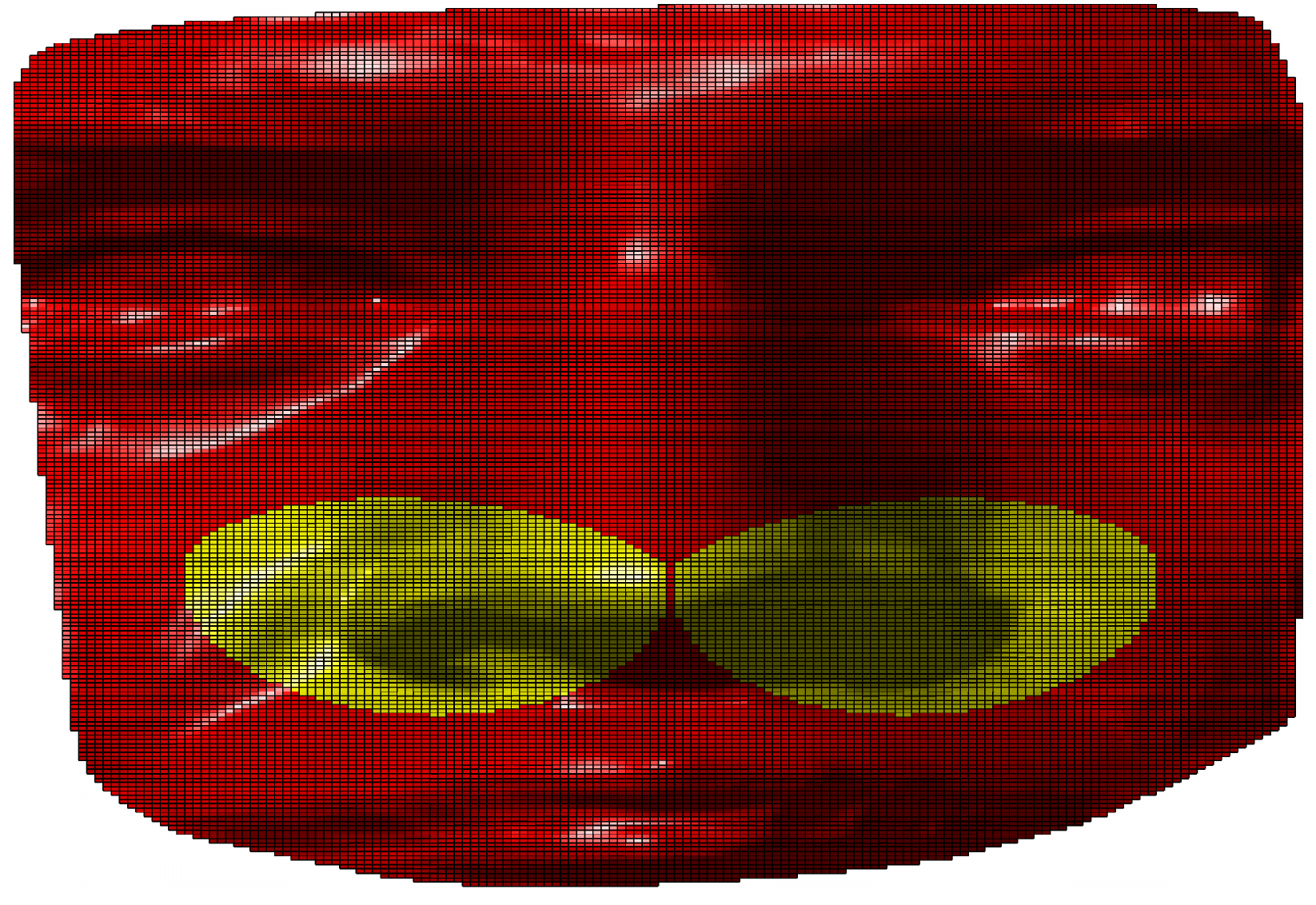}}
\subfloat[]{\includegraphics[width=0.2\textwidth]{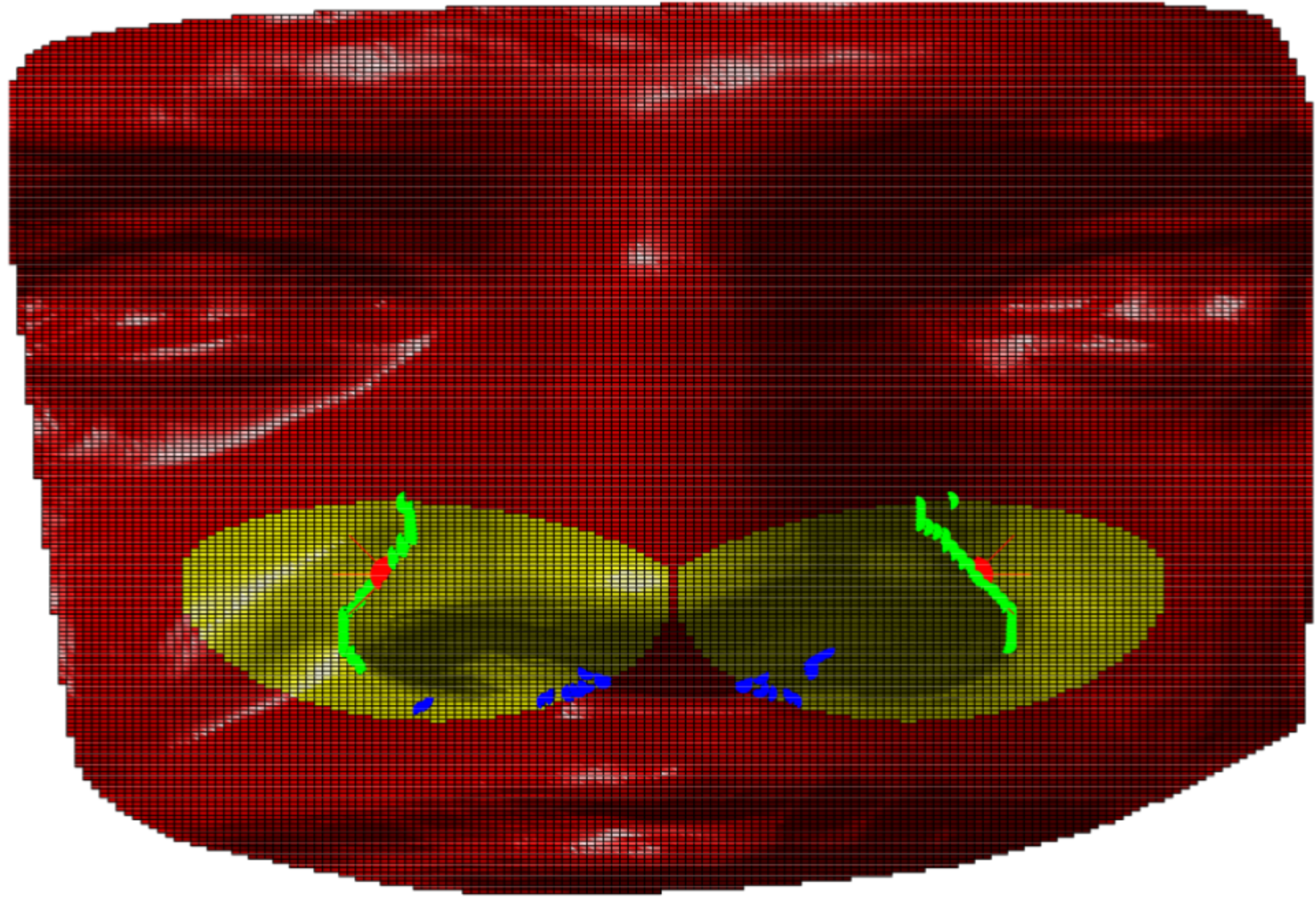}}\\
\subfloat[]{\includegraphics[width=0.2\textwidth]{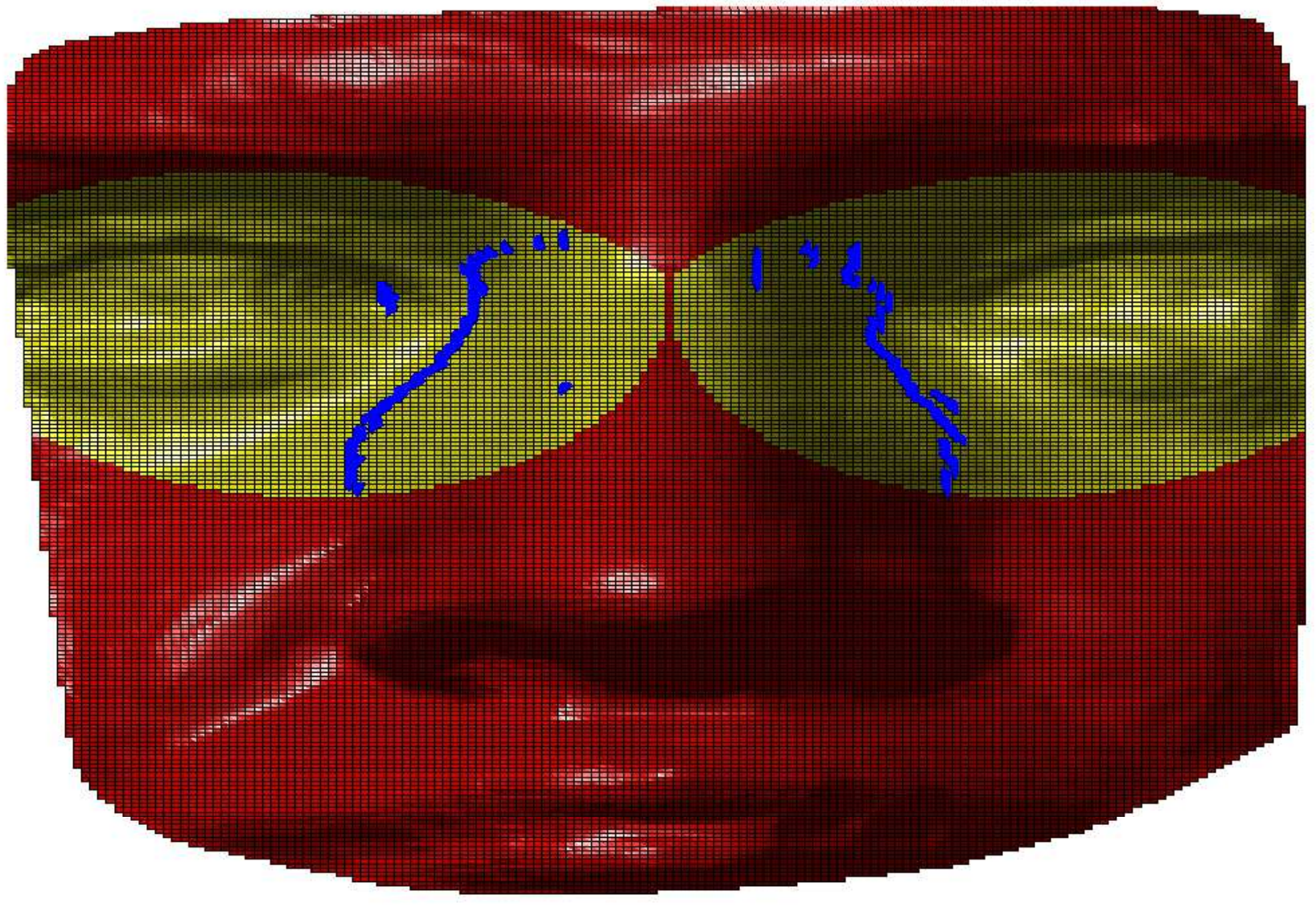}}
\subfloat[]{\includegraphics[width=0.2\textwidth]{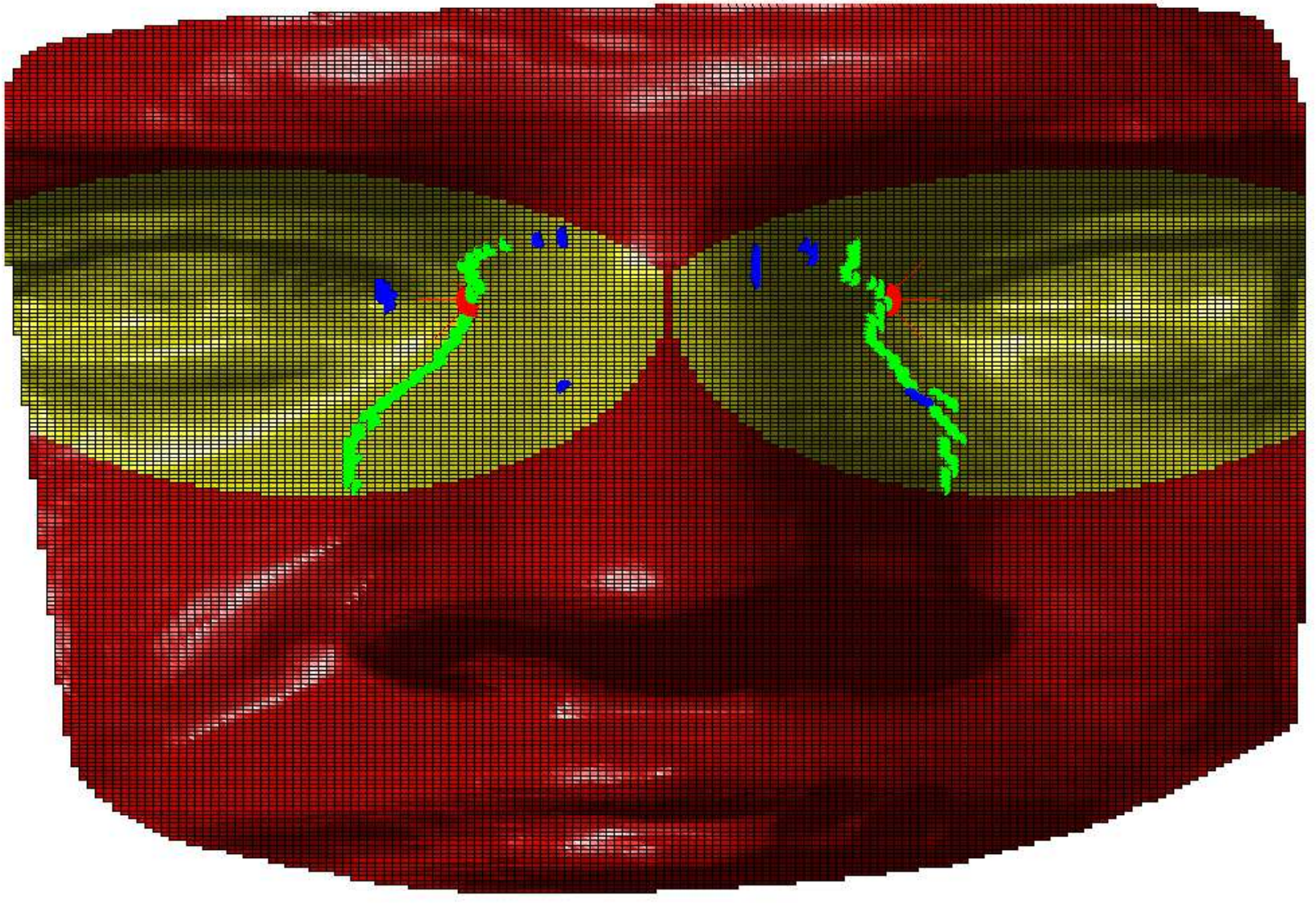}}
\caption{(a) and (b) RoIs for detection of the nasal alar groove and eye corners landmarks. (b) and (d) show green and blue points as the inliers and outliers, while the red points are the selected locations for {\bf{L3}}, {\bf{L6}}, {\bf{L2}} and {\bf{L7}}, respectively.}
\label{Fig:ROIALARROIEYE}
\end{figure}

\subsection{Nose alar groove and eye corners localisation}
The location of the nose tip ($\bf{L4}$) is moved to the origin and an RoI defined to detect the nasal alar grooves (Fig.~\ref{Fig:ROIALARROIEYE}-a) by,

\begin{equation}
	r = \left\{ \begin{array}{cc}
r_0 \cos^{a_1}(\theta) & 0\leq\theta<\pi \\
r_0 \cos^{a_2}(\theta) & \pi\leq\theta<2\pi \end{array}\right.
\label{eq:}
\end{equation}

\noindent where $r_0$, $a_1$ and $a_2$ are scalar constants determining the length and directivity of the lobes in the RoI. These are chosen to be able to crop the nasal alar region, while avoiding redundant parts (in subsequent experiments, $r_0$ = 30~mm, $a_1 = 4$ and $a_2 = 0.75$). $r$ and $\theta$ are the distance from the nose tip and angular rotation from the horizontal axis passing the nose tip location, respectively.
Similarly, the RoI used to detect the eye corners ($\bf{L2}$ and $\bf{L7}$) is depicted in Fig. \ref{Fig:ROIALARROIEYE}-c and is found using,

\begin{equation}
	r^\prime = \left\{ \begin{array}{cc}
r_0^{\prime} \cos^{a_3}(\theta^\prime) & 0\leq\theta^\prime<\pi \\
r_0^{\prime} \cos^{a_4}(\theta^\prime) & \pi\leq\theta^\prime<2\pi \end{array}\right.,
\label{eq:}
\end{equation}

\noindent in which $r_0^{\prime} = 45$~mm, $a_3 = 4$ and $a_4 = 0.75$. The polar coordinate system is characterised by $r^\prime$ and $\theta^\prime$, which are the distance from the nasal root ($\bf L1$) and the angular rotation from the horizontal axis passing $\bf L1$, respectively.

\begin{figure*}[!tb]
\centering
\includegraphics[width=0.7\textwidth]{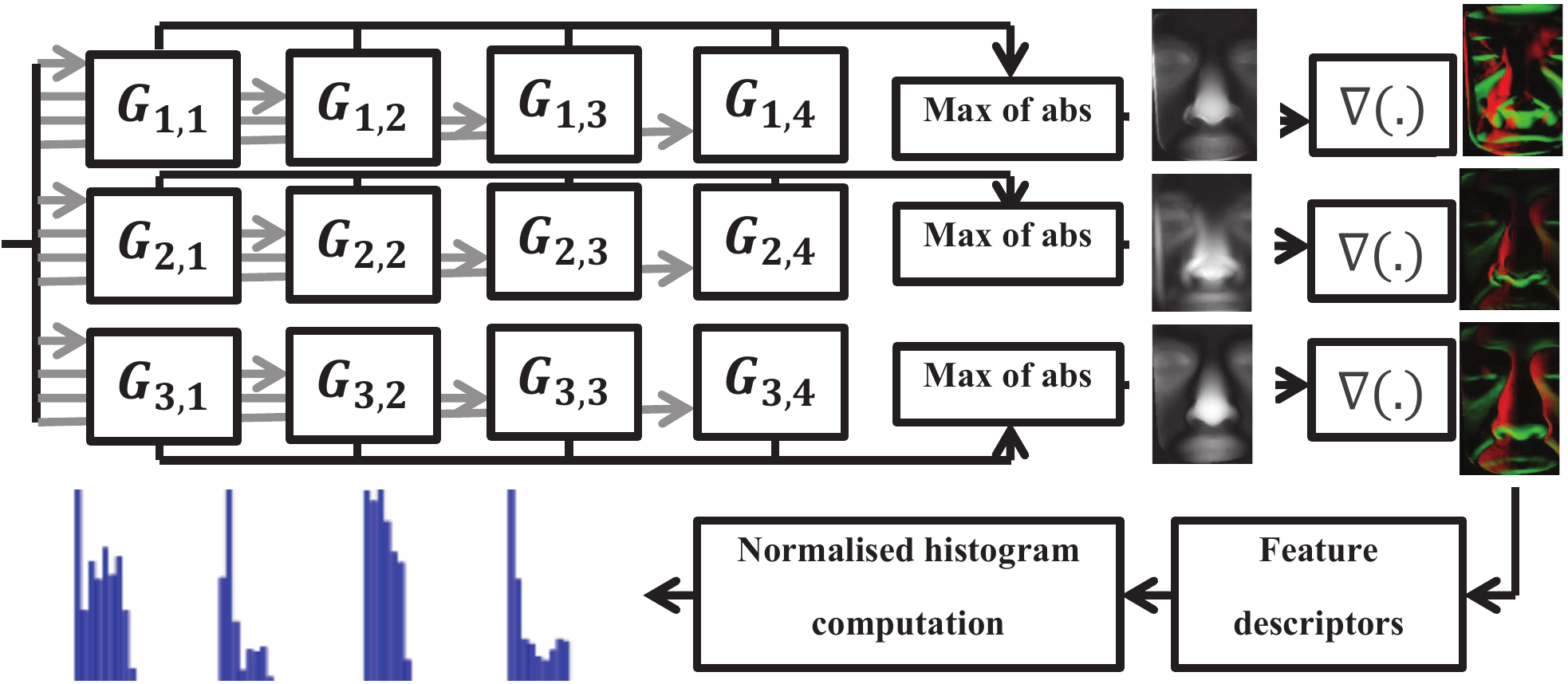}
\caption{The overall feature space creation procedure: 1) the wavelets are applied in different orientations and scales; 2) normals are computed on the maximum of absolute values of the filtered images per scale; 3) feature descriptors are applied; 4) normalised histograms are concatenated for all descriptors.}
\label{Fig:overallApp}
\end{figure*}

Planes parallel with the $xz$-plane are then intersected with each row of the RoIs. For the $j^{th}$ and $k^{th}$ intersections over the nasal alar groove and eye corner RoIs, curves ${\bf Q}^j$ and ${\bf Q^{\prime}}^k$ are found, respectively, and (\ref{eq:mindetector}) is used to find three minima for the nasal alar groove,

\begin{equation}
\left\{
\begin{array}{l}
	{{\bf{RMIN}}_j} = V_{3,\zeta}({{\bf{Q}}^j}_{\zeta}) \\
	{{\bf{LMIN}}_j} = V_{3,\eta}({{\bf{Q}}^j}_{\eta})
\end{array}\right.
\label{eq:zeta1}
\end{equation}

\noindent and the eye corners,

\begin{equation}
\left\{
\begin{array}{l}
	{{\bf{RMIN}}_k^\prime} = V_{3,\zeta}({{\bf{Q}}^{\prime k}}_{\zeta}) \\
	{{\bf{LMIN}}_k^\prime} = V_{3,\eta}({{\bf{Q}}^{\prime k}}_{\eta})
\end{array}\right.,
\label{eq:zeta2}
\end{equation}

\noindent where $\{ {{\bf{RMIN}}_j}, {{\bf{LMIN}}_j}\}$ and $\{{{\bf{RMIN}}_k^\prime}, {{\bf{LMIN}}_k^\prime}\}$ are matrices with three rows, in which each row has the location of the lowest minimum found from the $j^{th}$ and $k^{th}$ rows of the RoIs, for the right and left sides of $\bf{L4}$ and $\bf{L1}$, respectively. Then, for each row, ${{\bf{RMIN}}_j}$ and ${{\bf{LMIN}}_j}$ are compared and the pairs with the most similar Euclidean distances to the nose tip $\bf{L4}$ selected. Using a similar approach, the distance of ${{\bf{RMIN}}_k^\prime}$ and ${{\bf{LMIN}}_k^\prime}$ to $\bf{L1}$ for each row is computed and those pairs with the most similar distances kept. Figures \ref{Fig:ROIALARROIEYE}-b and -c illustrate these processes.

The points found as candidates for the nasal alar groove and eye corners might contain some outliers. This is because of the imaging noise and deformations on the face due to the facial expressions. To remove the outliers, an iterative approach is used. First, the {3D} Euclidean distances between the points on each consecutive row are computed. Then the standard deviation ($\sigma$) of the resulting vector is used to reject the points whose $\sigma$ is higher than a given threshold (in mm). This process continues until the number of inliers remains unchanged. Compared to the outlier removal method of \cite{Emambakhsh:2013}, which uses $K$-means clustering as a criterion to localise the outliers, this approach is deterministic and, unlike $K$-means, is not vulnerable to empty clusters. The outlier removal algorithm results in the green points labelled as the inliers in Fig. \ref{Fig:ROIALARROIEYE}-b and -d. The left and right pairs, which have the closest value of $y$ to that of the nose tip are selected as $\bf{L3}$ and $\bf{L6}$. Also, the points amongst the inliers in Fig. \ref{Fig:ROIALARROIEYE}-d, with the smallest depth values, are detected and the pair with the most similar distance to $\bf{L4}$ are selected as the eye corners ($\bf{L2}$ and $\bf{L7}$). The eye corners and nasal alar groove landmarks are the red points in Fig. \ref{Fig:ROIALARROIEYE}-b and -d, respectively.

\section{Feature extraction}
\label{sec:featextract}
The proposed feature space is based on surface normals. For an aligned depth map of the nasal region, represented by its point clouds as ${\bf{N}} = [{\bf{N_x}}, {\bf{N_y}}, {\bf{N_z}}]$ the normals are $\bf{n}  = [{\bf{n_x}}, {\bf{n_y}}, {\bf{n_z}}] = \nabla N$, where ${\bf{n_x}} \circ {\bf{n_x}} + {\bf{n_y}} \circ {\bf{n_y}} + {\bf{n_z}} \circ {\bf{n_z}} = {\bf{1}}$ ($\circ$ and $\bf 1$ represent the Hadamard product operator and matrix of ones, respectively). In order to reduce the sensitivity of the normal vectors to noise and enable the extraction of multi-resolution directional region-based information from the nasal region, instead of calculating the normal vectors directly from the nose surface, they are derived from the Gabor wavelet \cite{Tai:1996} filtered depth map. The algorithm proposed by Manjunath \emph{et al.} is used to minimise the wavelets overlap and redundancy in the filtered images \cite{Manjunath:1996}.

The discrete Fourier transform of the resampled Gabor wavelet ${\bf{G}}_{s, o}$ for the $s^{th}$ scale and $o^{th}$ orientation level ($s = \{1, 2, \ldots , s_m\}$ and $o = \{1, 2, \ldots , o_m\}$) is computed and its zero frequency component is set to zero. The Hadamard product of the resulting ${\bf{G}}^f_{s, o}$ and the Fourier transform of ${\bf{N_z}}$ is then calculated and the absolute value of its inverse Fourier transform is computed for each scale and orientation, i.e. ${\bf N^f_z}_{s, o}= \left|\mathcal{F}^{-1} \left\{ \mathcal{F}\{{\bf N_z}\} \circ {\bf{G}}^f_{s, o} \right\}\right|$. The maximum of all the corresponding elements of the filtered images is computed over all orientations for each scale $s$: $\left\{{\bf NGm}_s| \forall i,j,o: {\bf NGm}_s(i,j) \geq {\bf N^f_z}_{s, o}(i,j) \right\}$. In other words, ${\bf NGm}_s = \underset{o}{\text{max}}\left({\bf N^f_z}_{s, o}\right)$, where $\underset{o}{\text{max}}\left(.\right)$ computes the maximum of the corresponding elements along orientations $o$. Finally, the normal vectors of the resulting per scale maximal map ${\bf NGm}_s$ is calculated using the aligned nose coordinate maps ${\bf{N_x}}$ and ${\bf{N_y}}$,
\begin{equation}
\left\{
\begin{array}{l}
	{\bf{n}}_s = \nabla \left[{\bf{N_x}}, {\bf{N_y}}, {\bf NGm}_s \right] \\
	s = 1, 2, \ldots, s_m
\end{array}
\right.
\label{eq:nsGabor}
\end{equation}
\noindent where ${\bf{n}}_s=[{\bf{n_x}}_s, {\bf{n_y}}_s, {\bf{n_z}}_s]$ is a block matrix containing the normal vectors for the $s^{th}$ scale level.

\section{Localised feature descriptors using spherical patches and curves}
\label{sec:featdescription}
The feature descriptors are used to define a part of the nasal region, containing a set of normal vectors from the Gabor wavelets filters. Histograms of the resulting feature vectors for the $\bf X$, $\bf Y$ and $\bf Z$ maps are concatenated to create the feature space. This procedure is illustrated in Fig. \ref{Fig:overallApp} for $s_m = 3$ and $o_m = 4$. The feature descriptors are used to reduce the dimensionality of the feature space, decrease the redundancy and enable the use of probabilistic feature selection to lower the sensitivity to facial expressions while maintaining the most discriminative parts.

\begin{figure}[!tb]
\centering
\subfloat[]{\includegraphics[width=0.25\textwidth]{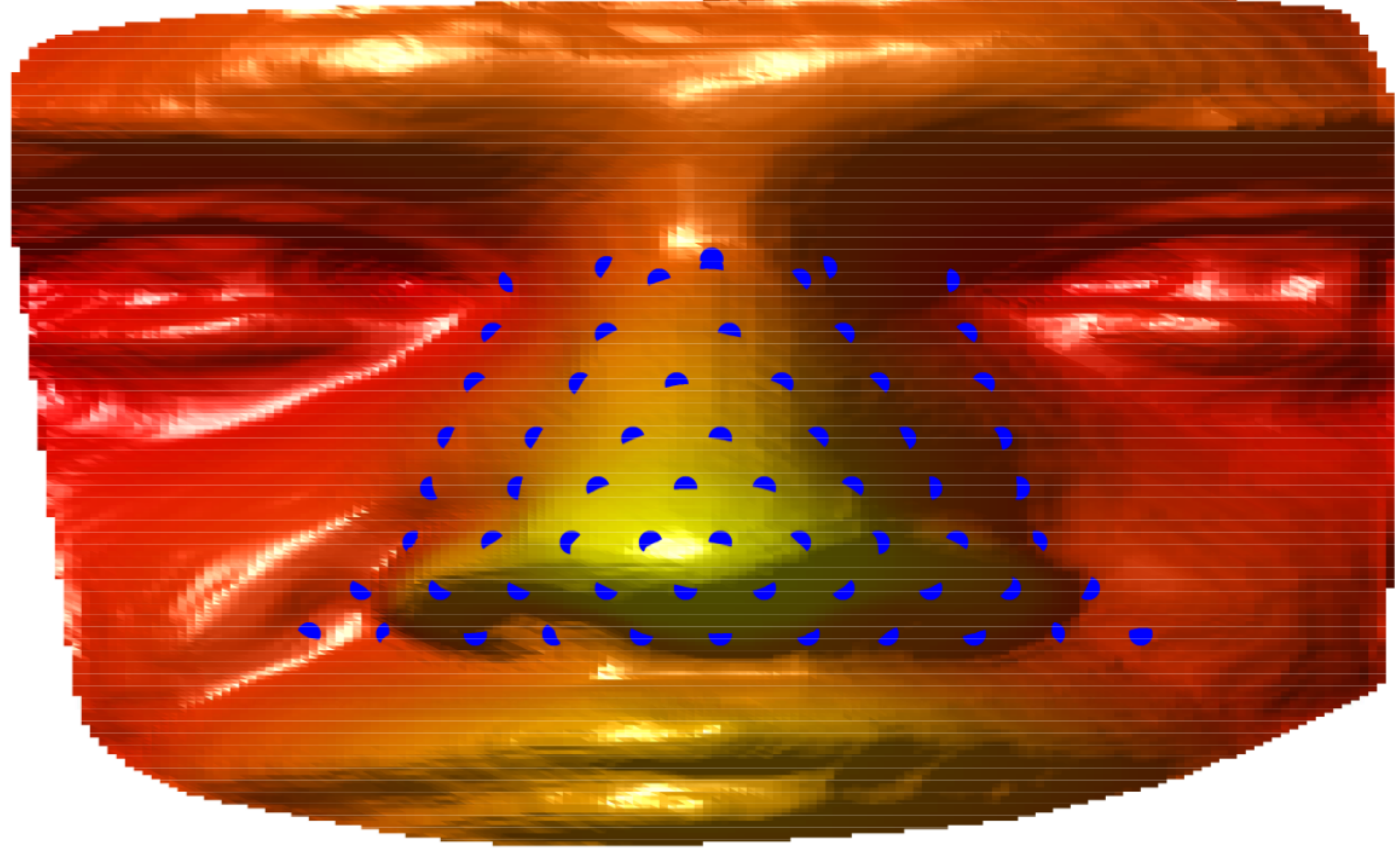}}
\subfloat[]{\includegraphics[width=0.25\textwidth]{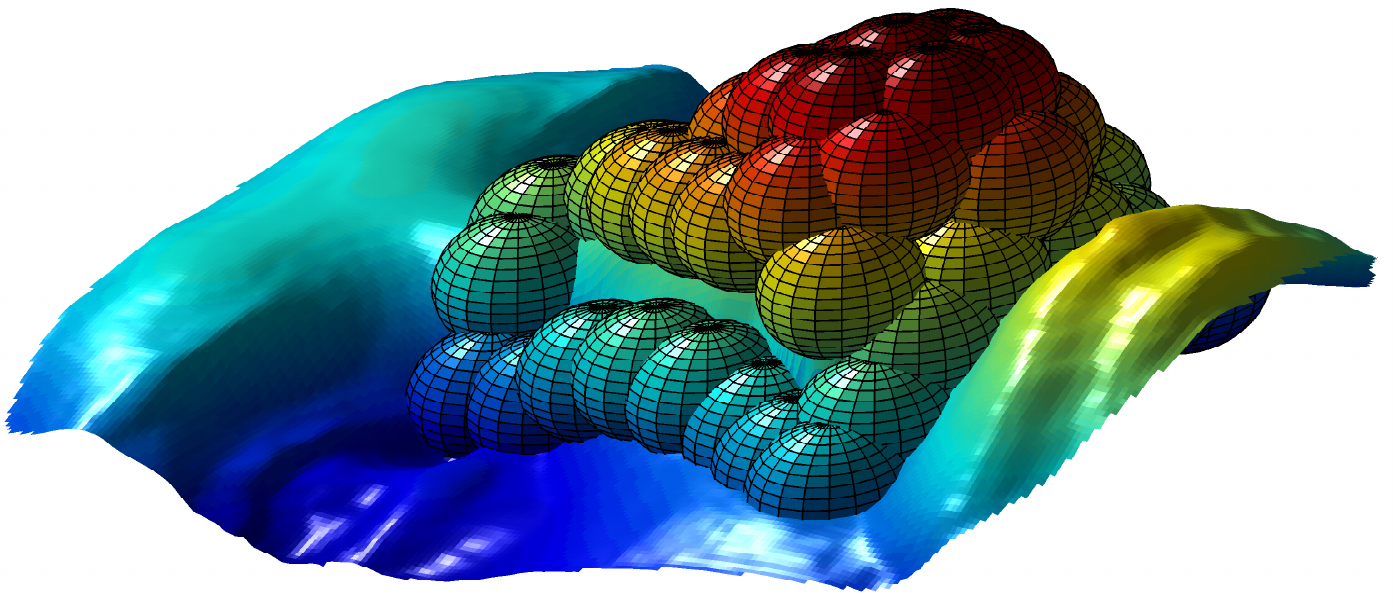}}\\
\subfloat[]{\includegraphics[width=0.25\textwidth]{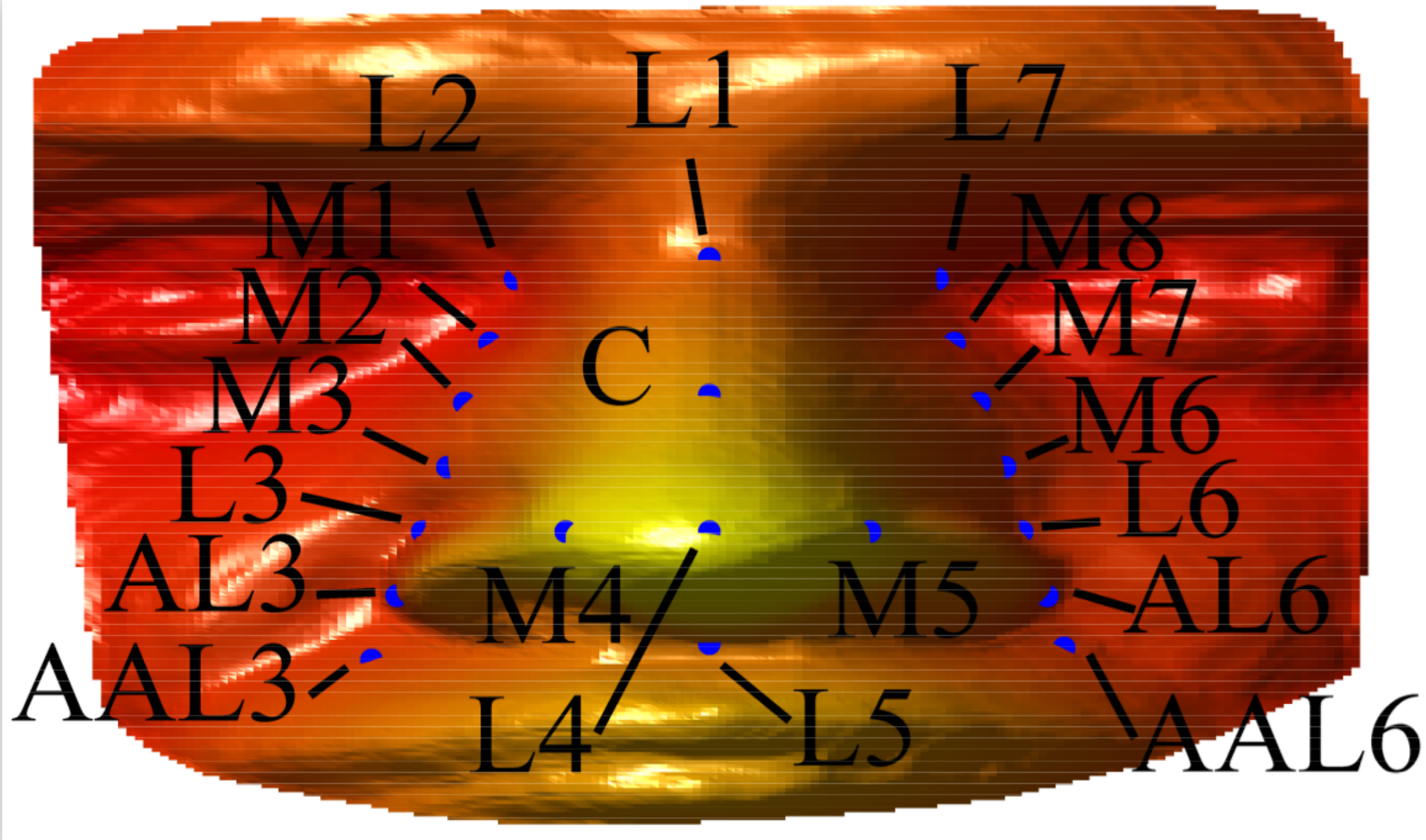}}
\subfloat[]{\includegraphics[width=0.25\textwidth]{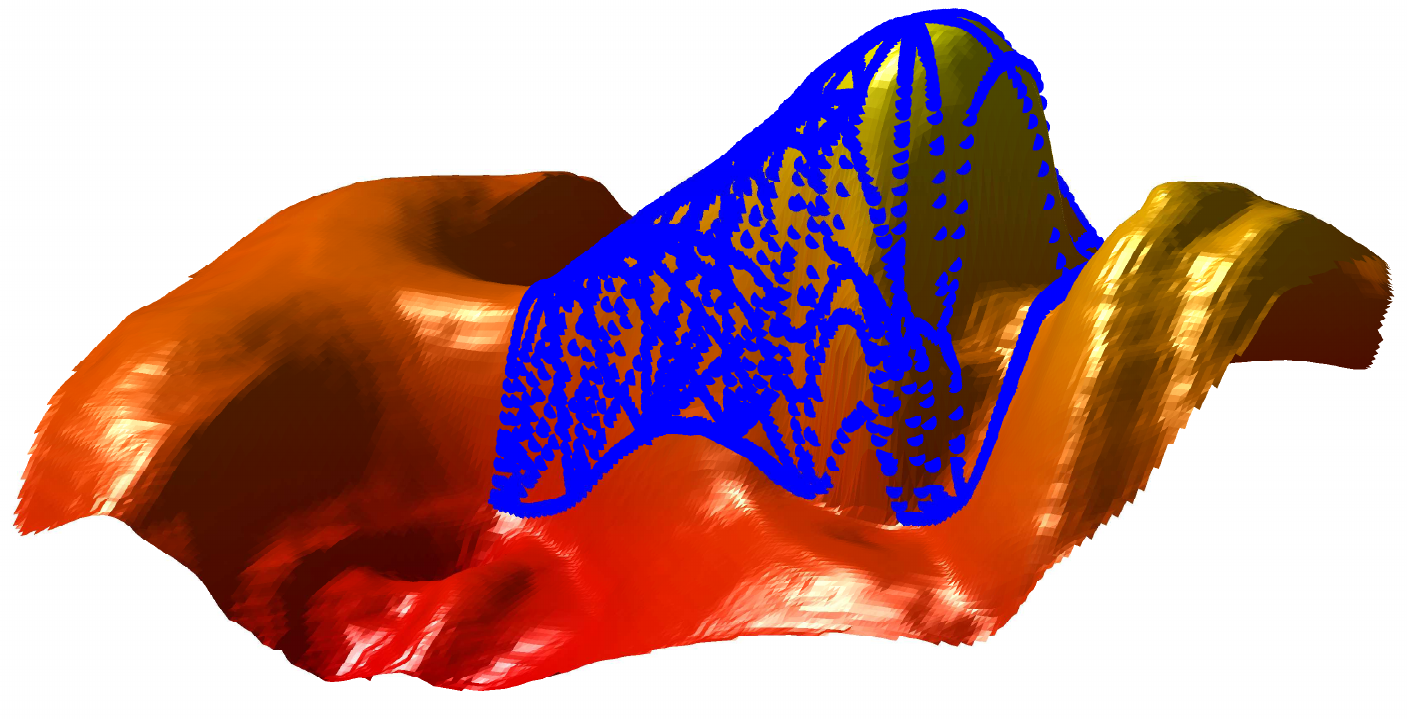}}
\caption{(a) Grid of landmarks used for the spherical patches in (b). The nasal curves (d) are found using the combination of new landmarks, illustrated in (c).}
\label{Fig:SphExp}
\end{figure}

The basic landmarks previously identified, see Fig. \ref{Fig:LandBlockLandmarksNames}-b, are used to create the new keypoints shown in Fig. \ref{Fig:SphExp}-a. These new landmarks are easily obtained by dividing the horizontal and vertical lines that connect the landmarks. A sphere centralised on each point is then intersected with the nasal surface and its inner parts are cropped. Then, the histogram of the normals of Gabor-wavelet filtered depth images are computed, based on the procedure explained in section~\ref{sec:featextract}. The intersection process is depicted in Fig.~\ref{Fig:SphExp}-b. A set of spheres of identical radii (in this case 7~mm) are intersected with the nose surface. These spherical feature descriptors provide the capability to evaluate the potential of overlapping spherical regions on the nasal surface, when used as feature vectors.

Alternatively, using different pairs of landmarks, a set of orthogonal planes to the nasal region can be found. Intersecting the planes with the nose surface results in a set of curves on the nasal region. For example, the normal vector of a plane passing through two nasal landmarks ${\bf{A_1}}$ and ${\bf{A_2}}$, and orthogonal to the $xy$ plane can be defined by $\hat{p}_{{\bf{A_1}}{\bf{A_2}}} = \hat{a}_z \times \frac{[{\bf{A_1}} - {\bf{A_2}}]}{\sqrt{[{\bf{A_1}} - {\bf{A_2}}]}}$, where $\hat{a}_z = [0, 0, 1]$ is the unit vector along the $z$-axis. When ${\bf{A_1}}$ and ${\bf{A_2}}$ are selected from the set of landmarks shown in Fig.~\ref{Fig:SphExp}-c, they can be used to create the set of curves shown in Fig.~\ref{Fig:SphExp}-d, which provide the feature descriptors. For each curve, the concatenated histograms of the $x$, $y$ and $z$ components of the normal vectors from the Gabor wavelet filters outputs are computed, giving the feature vector.

\section{Feature selection using GA}
\label{sec:featselection}
The feature selection step selects those subsets of feature vectors extracted from the curves and spherical patches that are more robust against facial expressions. For a given feature descriptor and $n$ different Gabor wavelets scales $\{s_1, s_2, \ldots, s_n\}$, the feature vector is computed by,
\begin{equation}
\left\{\begin{array}{l}
	{\bf{F}} = \left[{\bf{F}}_{s_1}, {\bf{F}}_{s_2}, \ldots, {\bf{F}}_{s_n} \right] ,\\
	{\bf{F}}_{s_k} = \left[{\bf{Fx}}_{s_k}, {\bf{Fy}}_{s_k}, {\bf{Fz}}_{s_k} \right],
\end{array}\right.
\end{equation}
\noindent where ${\bf{Fx}}_{s_k}$, ${\bf{Fy}}_{s_k}$ and ${\bf{Fz}}_{s_k}$ are the features of the $s_k^{th}$ scale, for the $x$, $y$ and $z$ surface normal components, respectively. For $K$ feature descriptors, each feature set of the normal maps is represented  by the concatenation of $K$ different histograms, of length $h_l$ from the feature descriptors, giving
\begin{equation}
\left\{
\begin{array}{l}
	{\bf{Fx}}_{s_k} = \left[{\bf{Hx}}_{1,s_k}, {\bf{Hx}}_{2,s_k}, \ldots, {\bf{Hx}}_{K,s_k}\right], \\
	{\bf{Fy}}_{s_k} = \left[{\bf{Hy}}_{1,s_k}, {\bf{Hy}}_{2,s_k}, \ldots, {\bf{Hy}}_{K,s_k}\right], \\
	{\bf{Fz}}_{s_k} = \left[{\bf{Hz}}_{1,s_k}, {\bf{Hz}}_{2,s_k}, \ldots, {\bf{Hz}}_{K,s_k}\right].
\end{array}
\right.
\label{eq:FXYZ}
\end{equation}
\noindent In (\ref{eq:FXYZ}), ${\bf{Hx}}_{i,s_k}$, ${\bf{Hy}}_{i,s_k}$ and ${\bf{Hz}}_{i,s_k}$ are the normalised histograms computed using the $i^{th}$ feature descriptor ($i=1,\ldots,K$) for the $s_k^{th}$ scale ($k=1,\ldots,n$) on the normal map ${\bf{n}}_{s_k}$, which is computed using (\ref{eq:nsGabor}).

Here the aim is to find a binary vector to be used as a switch to select the most robust and remove the vulnerable feature descriptors to facial expressions. Using a $1\times K$ binary vector ${\bf{Bn}}$, the vector ${\bf{B}}_{s_k}$, whose length is equal to the length of ${\bf{Fx}}_{s_k}$, ${\bf{Fy}}_{s_k}$ or ${\bf{Fz}}_{s_k}$ can be computed for the $s_k^{th}$ scale by,
\begin{equation}
\left\{
\begin{array}{l}
{\bf{B}}_{s_k} = \left[{\bf{B}}_{1,s_k}, {\bf{B}}_{2,s_k}, \ldots, {\bf{B}}_{K,s_k}\right],\\
{\bf{B}}_{i,s_k} = \left\{
\begin{array}{l}
\left[0, 0, \ldots , 0\right] \text{  if  } {\bf{Bn}}(i)=0 \\
\left[1, 1, \ldots , 1\right] \text{  if  } {\bf{Bn}}(i)=1
\end{array}\right..
\end{array}\right.
\label{eq:BnNeuc}
\end{equation}
\noindent The elements of ${\bf{B}}_{i,s_k}$ ($i=1,\ldots,K$) are set to zero or one, depending on the value of the $i^{th}$ element of ${\bf{{Bn}}}$. Finally, ${\bf{B}}_{s_k}$ is concatenated over all scales to create a binary vector $\bf B$, whose length is equal to the feature space dimensionality,
\begin{equation}
{\bf{B}} = \left[\underbrace{\overbrace{[{\bf{B}}_{s_1}, {\bf{B}}_{s_1}, {\bf{B}}_{s_1}]}^{\text{for all normals in scale 1}}, \ldots,\overbrace{[{\bf{B}}_{s_n}, {\bf{B}}_{s_n}, {\bf{B}}_{s_n}]}^{\text{for all normals in scale n}}}_{\text{length} = s_n \times 3 \times K \times h_l}\right].
\label{eq:FeatVectLength}
\end{equation}
\noindent The value of each element of $\bf B$ can be altered using the nucleus binary vector ${\bf{Bn}}$. A curve or patch is selected or omitted based on the value of ${\bf{Bn}}$ elements. If ${\bf{Bn}}(i)=1$, then the $i^{th}$ curve or patch is selected, otherwise it is omitted. By grouping the neutral samples for the gallery and the non-neutral samples for the test phase, and varying ${\bf{Bn}}$, the most expression robust curves and patches can be selected. As shown in (\ref{eq:BnNeuc}) and (\ref{eq:FeatVectLength}) when ${\bf{B}}_{s_1} = {\bf{B}}_{s_2} = \ldots = {\bf{B}}_{s_n}$, all curves and patches are selected or removed, simultaneously, for all scales. The resulting low dimensional samples are matched with those in the gallery using the Mahalanobis cosine distance,
\begin{equation}
{\bf{D_{g,p}}} = - \left( \frac{{\bf{X_{g}}}}{\sqrt{|{\bf{X_{g}}}\Sigma^{-1}{\bf{X_{g}}}^\intercal|}} \right) \Sigma^{-1} \left( \frac{{\bf{X_{p}}}}{\sqrt{|{\bf{X_{p}}} \Sigma^{-1} {\bf{X_{p}}}^\intercal|}} \right)^\intercal.
\label{eq:matching}
\end{equation}
\noindent The Kernel Fisher's analysis (KFA) algorithm with polynomial kernel is applied to the feature space to project the features to a lower dimensional space using a supervised approach. If $S_g$ and $S_p$ are the number of gallery and probe samples, respectively, and $d_p$ is the dimension of the projected subspace (in all subsequent experiments, $d_p=S_g$), ${\bf{X_{g}}}$ and ${\bf{X_{p}}}$ will be matrices of dimensions $S_g \times d$ and $S_p \times d$, respectively. $\Sigma$ is the $d_p \times d_p$ covariance matrix computed over ${\bf{X_{g}}}$, and ${\bf{D_{g,p}}}$ is a $S_g \times S_p$ distance matrix containing the matching errors. To maximise the probability of assigning the test samples ${\bf{X_{t}}}$ to their corresponding classes (subjects), when compared with the gallery samples ${\bf{X_{g}}}$, ${\bf{Bn}}$ can be varied and its optimum found by,
\begin{equation}
{{\bf{Bn}}}_{opt} = \operatorname*{arg\,max}_{{\bf{Bn}}} \{R_1 \}
\label{eq:OBJFUN}
\end{equation}
\noindent in which $R_1$ is the average probability that the label corresponding to the smallest matching error, found by (\ref{eq:matching}), is the same as the label of the probe sample. In other words, $R_1$ is the rank one recognition rate which is maximised as ${\bf{Bn}}$ is changed. The excellent capability of GA in high dimensional binary parameter space \cite{Queirolo:2010,Emambakhsh:2010} make it well suited for this non-convex optimisation problem. The GA used in this work is a modified Non-dominated Sorting Genetic Algorithm-II (NSGA-II) \cite{Deb:2002} which, in comparison with NSGA \cite{Srinivas:1994}, is an elitism-based approach, relies on an improved sorting algorithm, has lower computational complexity and does not require sharing parameter assignment \cite{Deb:2002}. The modified NSGA-II incorporates elitism over the individuals that increase the diversity of the population in addition to those with better fitness output. The parameter assignments for the GA are explained in section \ref{sec:paramassign}.

\renewcommand{\baselinestretch}{1.0}
\begin{table*}[!t]
\begin{center}
\begin{tabular}{| L{1.5cm} | L{1.75cm} | L{1.75cm} | L{1.75cm} | L{1.75cm} | L{1.75cm} | L{1.75cm} |}
\hline
	\bf Dataset & {\bf Nasal root} \newline \bf (L1) & {\bf{Left eye corner} \newline \bf (L2)} & {\bf{Left alar groove} \newline \bf (L3)} & \bf{Subnasale} \newline \bf (L5) & \bf{Right alar groove} \newline \bf (L6) & \bf{Right eye corner} \newline \bf (L7)\\ \hline \hline
	Bosphorus & 1.06 $\pm$ 0.58 & 1.76 $\pm$ 1.03 & 1.06 $\pm$ 0.62 & 1.11 $\pm$ 0.38 & 1.19 $\pm$ 0.60 & 2.12 $\pm$ 1.14\\ \hline
	FRGC & 2.04 $\pm$ 1.09 & 2.95 $\pm$ 1.61 & 1.29 $\pm$ 0.82 & 1.86 $\pm$ 0.85 & 1.22 $\pm$ 0.62 & 2.91 $\pm$ 1.53\\ \hline
\end{tabular}
\end{center}
\caption{Landmarking consistency error in mm.}
\label{Tab:consistency}
\end{table*}
\renewcommand{\baselinestretch}{2.0}

\section{Experimental results}
\label{sec:expresults}
\subsection{{3D} datasets and experiments}
Three datasets are used to evaluate the proposed recognition algorithms. The first one is the FRGC dataset, which is widely recognised as the largest {3D} face dataset, with 557 subjects. The captures in the dataset were obtained using the Minolta laser sensors over three different sets of sessions: Spring 2003, Fall 2003 and Spring 2004 \cite{Phillips:2005}. The samples in the Spring 2003 folders are known as the v1.0 \cite{Li:2014}, while the collection in the other two folders constitute v2.0. FRGC v1.0 and v2.0 have 267 subjects (838 samples) and 466 subjects (4007 samples),  respectively. To evaluate the algorithm on this dataset, three sets of experiments are defined. For the first set, FRGC v2.0 is divided into a 466 samples gallery and 3541 probe samples, an arrangement that has been extensively used in the literature \cite{Mian:2007,Osaimi:2009,Wang:2010,Li:2014,Berretti:2013,Mian:2008,Berretti:2010,Smeets:2013}. The second experiment is known is FRGC's ROC III on Exp III \cite{Phillips:2005}. This is a verification scenario, which uses the between season {3D} data samples. For this experiment, usually the equal error rate (EER) or 0.1\% false accept rate (FAR) is reported. The third evaluation using FRGC is termed expression vs. expression. FRGC consists of samples with neutral and non-neutral facial expressions, and using different sets of facial expressions for the probe samples enables neutral gallery vs. neutral probe and neutral gallery vs. non-neutral probe evaluations. The purpose of this experiment is to quantitatively evaluate how the performance of a face recognition system with a neutral gallery changes when the probe samples are replaced by non-neutral samples.

The other two datasets used are the Bosphorus {3D} face dataset \cite{Savran:2008} and the BU-3DFE dataset \cite{Yin:2007}, which contain captures of six prototypic expressions (anger, disgust, fear, happiness, sadness and surprise) in addition to neutral. BU-3DFE contains 100 subjects (56 female and 44 male) with age range 18 to 70 years. It contains four different levels of intensity of each facial expression and only one neutral sample per subject, making it one of the most challenging benchmarks for face recognition. For the Bosphorus database, one neutral sample per subject is used in the gallery (105 samples) and the remaining 2797 samples as probes \cite{Li:2014,Li:2011}. For both the Bosphorus and BU-3DFE a specific expression robustness evaluation uses the neutral expression as the gallery and the captures for each expression in turn as probes~\cite{Li:2011,Hajati:2012}.

\renewcommand{\baselinestretch}{1.0}
\begin{table*}[!t]
\begin{center}
\begin{tabular}{| L{1.75cm} | L{2.2cm} | L{2cm} | L{2cm} | L{1.7cm} | L{2cm} | L{2cm} |}
\hline
\bf Algorithm & \bf Threshold (mm) & \bf {\bf{L3}} & \bf {\bf{L6}} & \bf {\bf{L2}} & \bf {\bf{L7}} & \bf Nose tip {\bf{L4}} \\ \hline \hline
Proposed \newline method & $<$ 10 \newline $<$ 12 \newline $<$ 15 \newline $<$ 20 & 99.55\% \newline 99.62\% \newline 99.66\% \newline 99.69\% & 99.35\% \newline 99.62\% \newline 99.66\% \newline 99.66\% & 96.69\% \newline 97.73\% \newline 99.04\% \newline 99.59\% & 94.59\% \newline 96.56\% \newline 98.38\% \newline 99.62\% & 97.52\% \newline 99.04\% \newline 99.66\% \newline 99.79\% \\ \hline
Creusot \emph{et al.} \cite{Creusot:2013} & $<$ 10 \newline $<$ 12 \newline $<$ 15 \newline $<$ 20 & 97.96\% \newline 99.18\% \newline 99.82\% \newline 99.90\% & 98.43\% \newline 99.71\% \newline 99.86\% \newline 99.90\% & 98.82\% \newline 99.65\% \newline 99.93\% \newline 99.93\% & 98.50\% \newline 99.43\% \newline 99.75\% \newline 99.86\% & 95.47\% \newline 98.15\% \newline 98.97\% \newline 99.33\% \\ \hline
\end{tabular}
\end{center}
\caption{Landmarking precision accuracy from the ground truth over the Bosphorus dataset samples.}
\label{Tab:LandmarkingComp}
\end{table*}
\renewcommand{\baselinestretch}{2.0}

\renewcommand{\baselinestretch}{1.0}
\begin{figure}[!tb]
\centering
\includegraphics[width=0.5\textwidth]{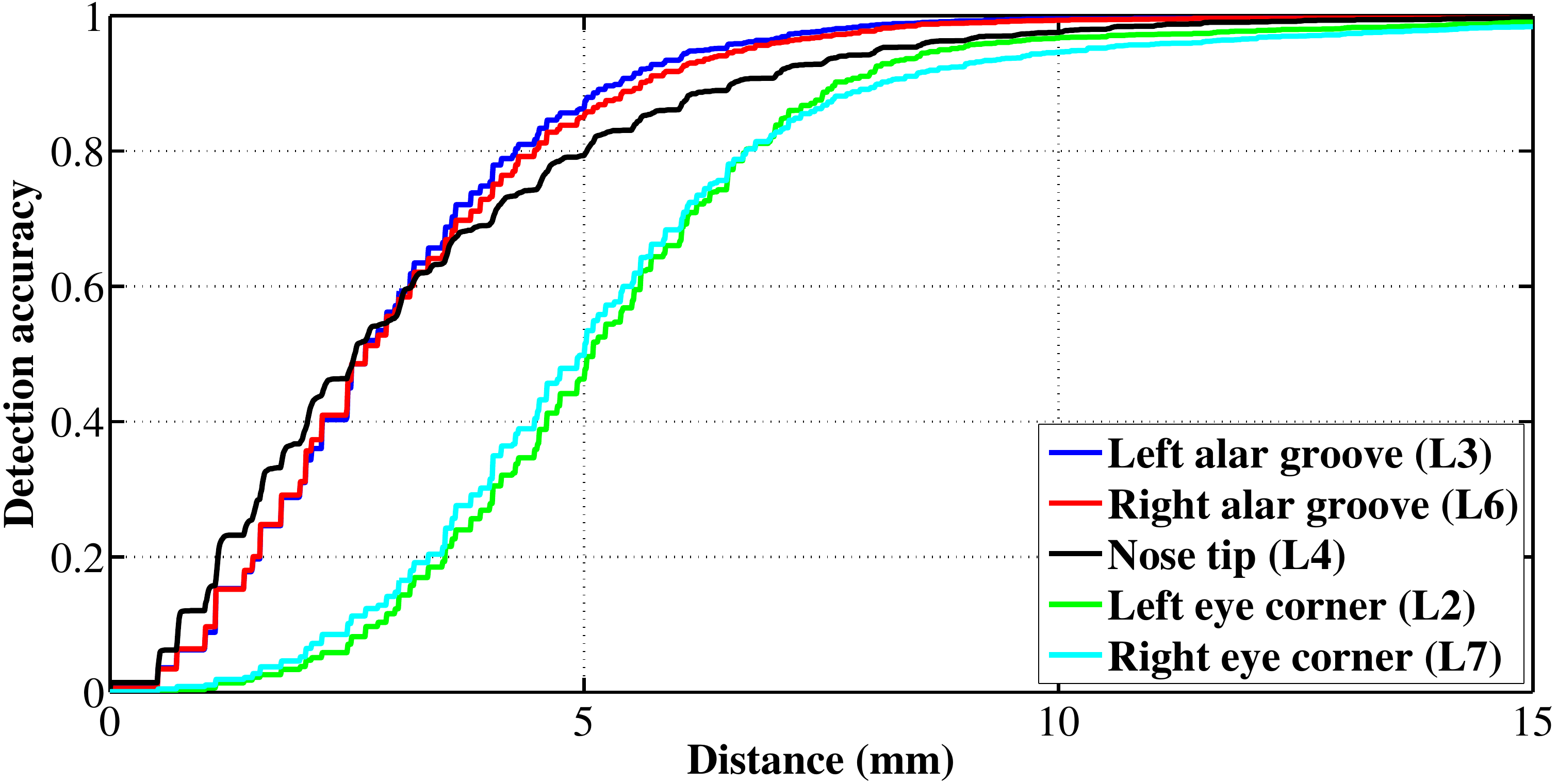}
\caption{Precision curves for the proposed landmarking algorithm computed using Bosphorus.}
\label{Fig:precision}
\end{figure}
\renewcommand{\baselinestretch}{2.0}

\subsection{Landmarking consistency and accuracy}
\label{Subsec:LandConsAcc}
The face recognition rates reported in the following subsections provide indirect evidence of the landmarking algorithm's consistency. However, to provide an independent assessment, the within-class similarity of the landmarking is investigated. In this evaluation, all subjects in the datasets are translated so that the nose tip is located at the origin. Then, the mean of the positions of each landmark for each subject's different facial expressions is computed. This process is performed for all subjects in the dataset and the averages and standard deviations are calculated.

For an ideal landmarking algorithm, the average for each landmark would be zero. However, due to the noise in the data and image acquisition errors, in practice the averages are non-zero. Although FRGC contains 557 subjects, some only have one sample and these are discarded, as it is impossible to compute the mean distance for such subjects. 

Table \ref{Tab:consistency} shows the results for the FRGC and Bosphorus datasets, when the rotation angles in (\ref{eq:forsaddlecand}) for the nasal root and $\bf{S}_{\phi}$ for the subnasale detection are ${\max}_m (\gamma_m) = \beta = \phi = \pi/3$ and the angles in (\ref{eq:zeta1}) and (\ref{eq:zeta2}) for the eye corners and nasal alar groove landmarks are $\eta = -\zeta = \pi/4$. The errors are higher for the FRGC dataset as its samples are noisier, especially those in the Spring 2003 folder. The most consistent pair of landmarks on both datasets are the two nasal alar ({\bf{L3}} and {\bf{L6}}), while the errors are slightly higher for the eye corners ({\bf{L2}} and {\bf{L7}}). The location of subnasale ({\bf{L5}}) is more consistently detected for the samples of the Bosphorus dataset. This is mainly due to the {3D} reconstruction noise for the higher frequency regions, as is in subnasale, for the FRGC samples. The other important factor for a landmarking algorithm is its accuracy and, in order to evaluate this, the locations of the landmarks are compared with the ground truth provided by the Bosphorus dataset. First, the PCA alignment matrix and $\theta_z^{opt}$ are applied to the ground truth landmarks to remap the points to the aligned faces domain. The precision curve \cite{Raguram:2011} is then computed for the landmarks $\{{\bf{L2}}, {\bf{L3}}, {\bf{L4}}, {\bf{L6}}, {\bf{L7}}\}$. As the location of nasal root ($\bf{L1}$) has not been assigned by the dataset providers, it is excluded from the accuracy evaluations. The curves shown in Fig. \ref{Fig:precision} are found over the action units samples (2150 observations) and samples with neutral and non-neutral expressions in the Bosphorus dataset (653 observations), constituting 2803 samples. For the range of $>10$~mm, all the landmarks are detected with $\geq 95\%$ accuracy. A comparison with the landmarking results reported by Creusot \emph{et al.} \cite{Creusot:2013} is shown in Table ~\ref{Tab:LandmarkingComp}. Although the results in \cite{Creusot:2013} are provided for different types of facial expressions, considering the number of samples and the recognition rates, they can be computed for all three action units, neutral and non-neutral samples in the Bosphorus dataset. For the nasal alar and tip, the proposed algorithm has higher accuracy, in particular the accuracy for the nose tip is improved by 2\%. However, for the eye corners, the algorithm in \cite{Creusot:2013} performs 2\% better. Although the proposed landmarking algorithm is not as robust as the approach of \cite{Creusot:2013} in the cases of partial and self-occlusions, it has the advantage of not requiring a training step.

\subsection{Robustness against facial expressions}
\label{sec:paramassign}
This section explains the parameters used for the feature descriptors, Gabor wavelets, the KFA and the GA optimiser. The supplementary material presents more extensive experiments that evaluate the effects of varying these parameters on the overall face recognition performance.

{\bf{Feature descriptors and Gabor wavelets}}: The radius of the spherical patches is 11~mm, while the numbers of histogram bins are 21 and 15 for the spherical and nasal curves, respectively. The Gabor wavelets at each orientation and scale can be defined by the lower and higher frequency levels ($\Omega_l$ and $\Omega_h$) and the maximum number of orientation and scale levels ($o_m$ and $s_m$). The parameter values used to obtain the results in the subsequent sections are: $\Omega_l=0.05$, $\Omega_h=0.7$, and $o_m = s_m = 4$.
For the Gabor wavelet implementation the ``Feature Extraction and Gabor Filtering'' code is employed\footnote{\url{http://old.vision.ece.ucsb.edu/texture/software/}}.
The supplementary material includes extensive results of the effects of the landmarking distribution for the spherical patches and Gabor wavelets parameters on the face recognition ranks.

{\bf{KFA}}: The polynomial kernel $\left({\bf{X_g^r}}{{\bf{X_p^r}}^\intercal} + {k_1} \right)^{\circ k_2}$ is applied to the $S_g \times d_r$ and $S_p \times d_r$ gallery and probe feature matrices $\bf{X_g^r}$ and $\bf{X_p^r}$, respectively, where $(.)^\circ$ is the Hadamard element-wise power operator. $d_r$ is the input feature space dimensionality, which is reduced to $d$ after KFA is utilised ($d = S_g - 1$ for all experiments in this paper). $k_1$ and $k_2$ are the scalar parameters of the kernel, set to 0 and 2.65, respectively. The KFA implementation is based on a modified version of the publicly available "Pretty Helpful Development" functions for face recognition toolbox (PHD toolbox)\footnote{luks.fe.uni-lj.si/sl/osebje/vitomir/face\%5Ftools/PhDface/}.

{\bf{GA}}: The modified NSGA-II algorithm used to perform the global optimisation to maximise (\ref{eq:OBJFUN}) stops when the variation in the stalled best fitness value is $<10^{-4}$, which is achieved after approximately 70,000 iterations for the curves and spherical patches. The population size is 15 times the number of variables, while the Pareto and Cross-over and Migration fractions are 0.35, 0.8 and 0.2, respectively. The uniform creation function is used to initialise the population and the phenotype distance crowding measurement is utilised to compute the individuals' distance measure. The code is implemented using Matlab's "Global Optimization Toolbox"\footnote{\url{http://mathworks.com/help/gads/gamultiobj.html}}.

\renewcommand{\baselinestretch}{1.0}
\begin{table*}[!t]
\begin{center}
\begin{tabular}{| L{2.5cm} | C{1.25cm} | C{1.25cm} | C{1.25cm} | C{1.25cm} | C{1.25cm} | C{1.25cm} | C{1.25cm} |}
\hline
\multirow{2}{*}{\bf Feature} & \multicolumn{7}{c|}{\bf Number of gallery samples per subject (No. gallery samples/ No. probe samples)} \\ \cline{2-8}
	\bf descriptors & \bf 1 \newline (482/4330) & \bf 2 \newline (880/3848) & \bf 3 \newline (1206/3408) & \bf 4 \newline (1432/3006) & \bf 5 \newline (1610/2648) & \bf 6 \newline (1752/2326) & \bf 7 \newline (1757/2034) \\ \hline \hline
{{Spherical patches}} & 96.2\% & 98.9\% & 99.4\% & 99.6\% & 99.6\% & 99.7\% & 99.75\% \\ \hline
{{Curves}} & 91.6\% & 96.8\% & 98.1\% & 98.8\% & 98.9\% & 99.3\% & 99.4\% \\ \hline
    \end{tabular}
\end{center}
\caption{R$_1$RR performance for varying the training size per subject, when all samples of the FRGC dataset are merged from the three seasons.}
\label{Tab:FRGCTrSize}
\end{table*}
\renewcommand{\baselinestretch}{2.0}

\subsubsection{Feature selection}
\begin{figure}[!t]
\centering
\includegraphics[width=0.5\textwidth]{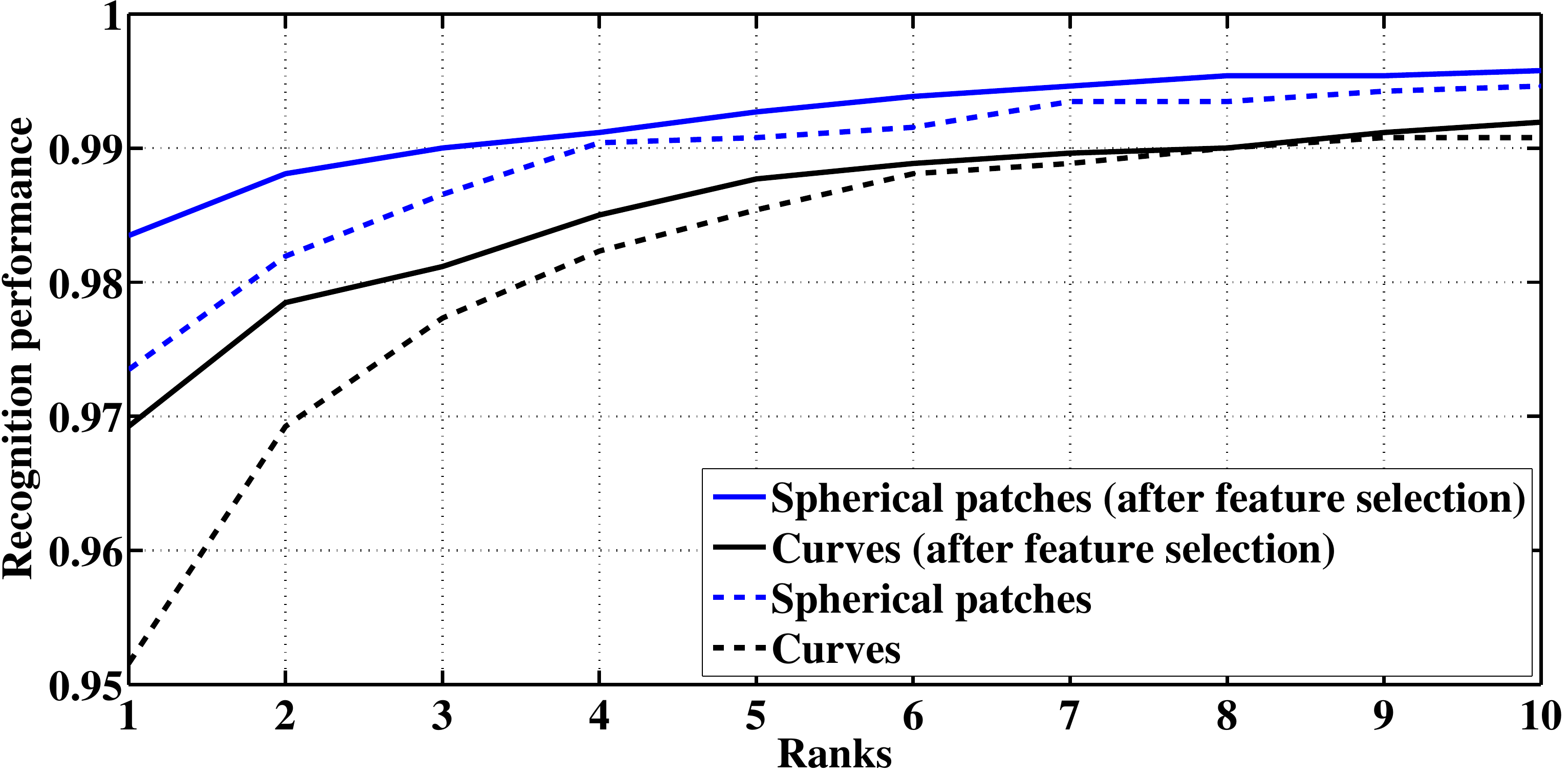}
\caption{CMC curves before and after feature selection on Bosphorus: neutral gallery vs. non-neutral probe.}
\label{Fig:BeforeAfterRanksCMCcurves}
\end{figure}

The algorithm described in section \ref{sec:featselection} is used to select the most discriminative patches and curves. For the feature selection stage, the neutral samples are employed for training and all the non-neutral samples for the test phase. The polynomial kernel is used for subspace projection. Then, the Mahalanobis cosine distance of (\ref{eq:matching}) is applied for the matching step at each iteration of the GA. 
In order to quantitatively illustrate how a subset of selected features can boost the recognition ranks, as an example, Fig.~\ref{Fig:BeforeAfterRanksCMCcurves} shows the face recognition ranks before and after applying the feature selection over the Bosphorus dataset. Over the first two ranks, the average improvement for both curves and patches is $\approx$1\% (more extensive results for the feature selection and feature space parameters are provided in supplementary material).

\renewcommand{\baselinestretch}{1.0}
\renewcommand{\baselinestretch}{2.0}

For all experiments presented in the following sections, different datasets are used for feature selection and testing, such that the test dataset is always completely unseen by the feature selector. Either the Bosphorus or the BU-3DFE dataset is used for feature selection, as these both have high variations in their expression types. The Bosphorus dataset is used for feature selection for all experiments in which FRGC or BU-3DFE are the test datasets, while BU-3DFE is used for feature selection with the Bosphorus dataset.


\subsubsection{Variable training size for FRGC}
To investigate the effects of using a different number of training samples per subject, the training size is increased from 1 to 7 for each subject in the FRGC dataset and the R$_1$RR computed for the selected patches and curves, see Table \ref{Tab:FRGCTrSize}. In this experiment, all the folders in FRGC from different seasons are merged. Then, for each subject, the number of samples in the gallery is changed and the average R$_1$RR are reported after the samples are interchanged in the gallery and probe. The results show the high discrimination of the feature space for the spherical patches over the FRGC dataset. For example, when only one sample per subject is used in the gallery (482 gallery samples vs. 4330 probe samples to recognise 482 subjects), a R$_1$RR of 96.2\% is achieved. To the best of our knowledge, this is the highest {3D} nasal region recognition rank ever obtained from this dataset for a single training sample and comparable with many state-of-the-art {3D} face recognition algorithms, which use the whole facial domain. Although the curves have lower R$_1$RR for one training sample per subject, there is a big increase in their recognition performance when the samples per subject are increased. For instance, when 2 samples per subject are used in the gallery, the nasal curves R$_1$RR performance increases by $>5\%$.

\subsubsection{FRGC v2.0 and ROC III}
\begin{figure}[!t]
\centering
\subfloat[]{\includegraphics[width=0.5\textwidth]{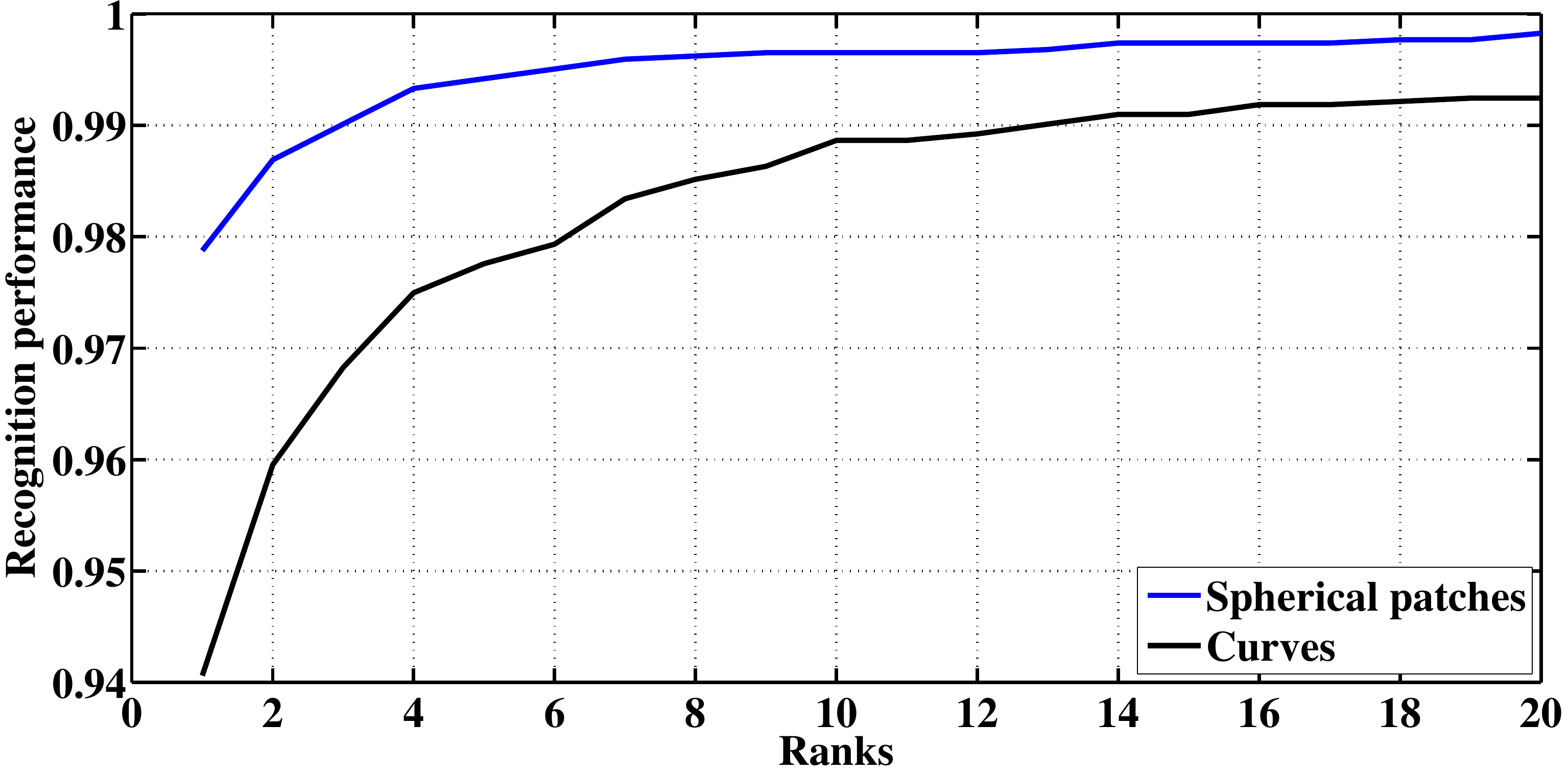}}\\
\subfloat[]{\includegraphics[width=0.5\textwidth]{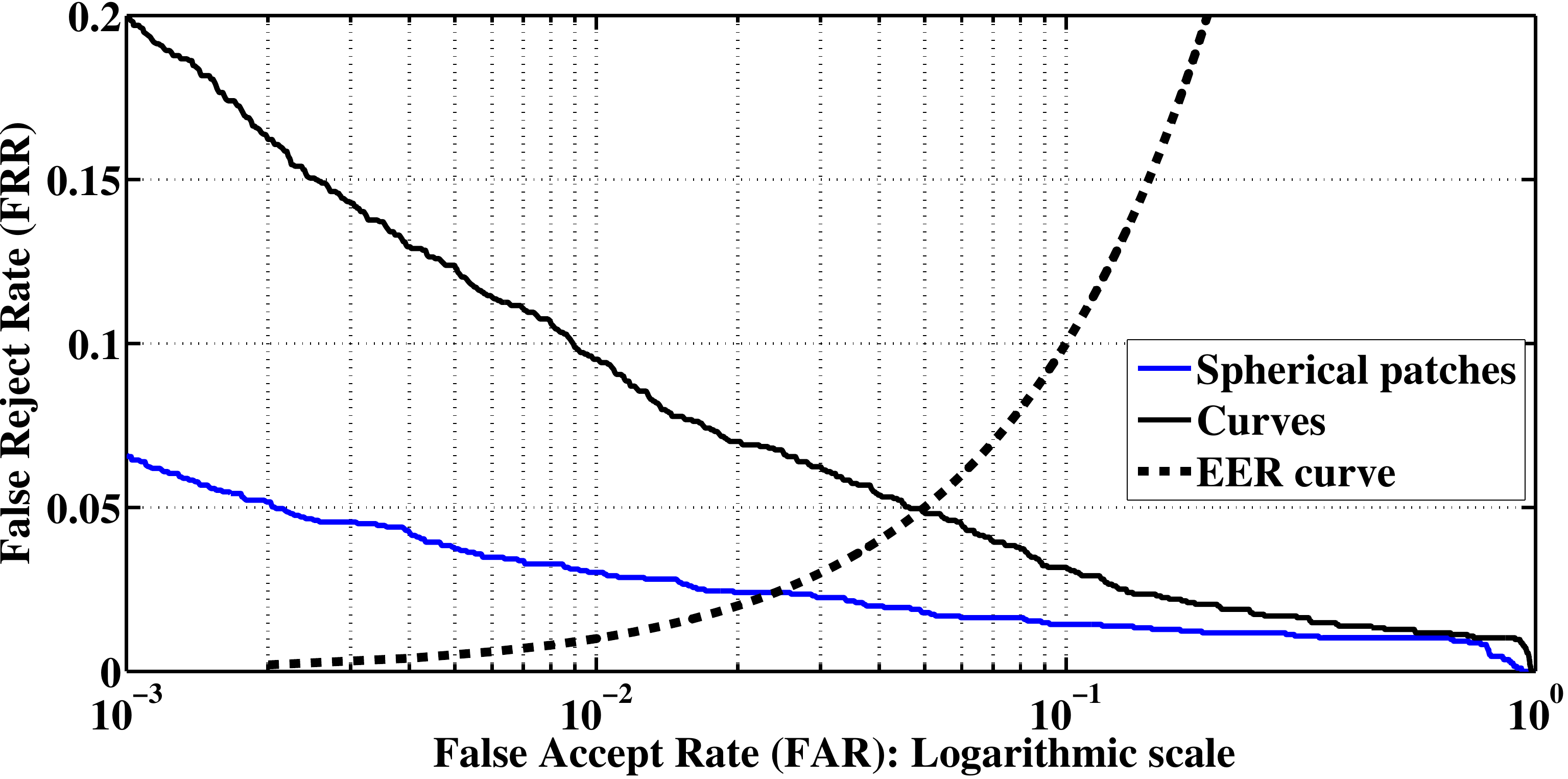}}
\caption{(a) CMC curves for FRGC v2.0; (b) Between seasons verification results for FRGC: ROC III.}
\label{Fig:FRGCCMCROC}
\end{figure}

The recognition performance for the FRGC v2.0 dataset and the ROC III experiment are widely used face recognition benchmarks. The dataset contains samples with different facial expressions for both the gallery and probe. The rank recognition performance shown in Fig.~\ref{Fig:FRGCCMCROC}-a increases with the rank to~$>99\%$ for the spherical patches. A similar trend exists for the cumulative matching characteristic (CMC) curves increasing with a high gradient for all ranks $>1$.

\renewcommand{\baselinestretch}{1.0}
\renewcommand{\baselinestretch}{2.0}

\renewcommand{\baselinestretch}{1.0}
\begin{table*}[!tb]
\centering
\begin{center}
\begin{tabular}{| L{2.6cm} | L{1.6cm} |  L{1.5cm} | L{1.9cm} |  L{2.1cm} | L{2.1cm} |  L{2cm} | }
    \hline
	\scriptsize \bf Algorithm & \scriptsize \bf Modality & \scriptsize \bf Rank-one \newline FRGC v2.0 & \scriptsize \bf EER ROC III & \scriptsize \bf 0.1\% FAR \newline ROC III & \scriptsize \bf Neutral vs. \newline Neutral & \scriptsize \bf Neutral vs. \newline Non-neutral \\ \hline \hline
\scriptsize {{Spherical patches}} \newline {{Curves}} & \scriptsize {3D} Nose & \scriptsize {97.9\%} \newline 94.1\% & \scriptsize 2.4\% \newline 4.9\% & \scriptsize 93.5\% \newline 80.0\% & \scriptsize {98.45\%} (R$_1$RR) \newline 95.8\% (R$_1$RR) & \scriptsize {{98.5\%}} (R$_1$RR) \newline 97.5\% (R$_1$RR) \\  \hline
\scriptsize Smeets \emph{et al.} \cite{Smeets:2013} & \scriptsize {3D} Face & \scriptsize 89.6\% & \scriptsize 3.8\% & \scriptsize 77.2\% &  - & - \\ \hline
\scriptsize Osaimi \emph{et al.} \cite{Osaimi:2009} & \scriptsize {3D} Face & \scriptsize 96.5\% &  - & \scriptsize 94.05\% & \scriptsize 98.35\% (0.1\% FAR) & \scriptsize 97.8\% (0.1\% FAR) \\ \hline
\scriptsize Spreeuwers \emph{et al.} \cite{Spreeuwers:2011} & \scriptsize {3D} Face \newline {3D} Nose & \scriptsize {{99.0\%}} \newline 94.5\% &  - & \scriptsize 94.6\% \newline 83.7\% &  - & - \\ \hline
\scriptsize Drira \emph{et al.} (2013) \cite{Drira:2013} & \scriptsize {3D} Face & \scriptsize 97.0\% &  - & \scriptsize 97.1\% & \scriptsize 99.2\% (R$_1$RR) & \scriptsize 96.8\% (R$_1$RR) \\ \hline
\scriptsize Aly\"uz \emph{et al.} (2010) \cite{Alyuz:2010} & \scriptsize {3D} Face \newline {3D} Nose & \scriptsize 97.5\% \newline 91.81 & \scriptsize 1.91\% \newline - & \scriptsize 85.6\% \newline - & \scriptsize 98.39\% (R$_1$RR) \newline - & \scriptsize 96.40\% (R$_1$RR) \newline - \\ \hline
\scriptsize Wang \emph{et al.} (2010) \cite{Wang:2010} & \scriptsize {3D} Face & \scriptsize 98.39\% & - & \scriptsize 98.04\% & \scriptsize 99.2\% (0.1\% FAR) & \scriptsize 97.7\% (0.1\% FAR) \\ \hline
\scriptsize Wang \emph{et al.} (2008)  \cite{Wang:20082} & \scriptsize {3D} Nose & \scriptsize 95\% (44mm) \newline 78\% (24mm) & - & - & - & - \\ \hline
\scriptsize Drira \emph{et al.} (2009) \cite{Drira:2009} & \scriptsize {3D} Face/Nose \newline (125 gallery) \newline (125 probe) & \scriptsize 88\% (Face) \newline 77.5\% (Nose) & - & - & - & - \\ \hline
\scriptsize Chang \emph{et al.} \cite{Chang:2006} & \scriptsize {3D} Nose & - & \scriptsize Neutral 12\% \newline Non-neutral 23\% & - & \scriptsize 97.1\% (R$_1$RR) & \scriptsize 86.1\% (R$_1$RR) \\ \hline
\scriptsize Emambakhsh \emph{et al.} \cite{Emambakhsh:2013} & \scriptsize {3D} Nose & \scriptsize 89.61\% & \scriptsize Neutral 8\% \newline Non-neutral 18\% & - & \scriptsize 90.87\% (R$_1$RR) & \scriptsize 81.61\% (R$_1$RR) \\ \hline
\scriptsize Li \emph{et al.} (2014) \cite{Li:2014} & \scriptsize {3D} Face & \scriptsize 96.3\% & - & - & \scriptsize 98.0\% (R$_1$RR) & \scriptsize 94.2\% (R$_1$RR) \\ \hline
\scriptsize Queirolo \emph{et al.} \cite{Queirolo:2010} & \scriptsize {3D} Face & \scriptsize {{99.6\%}} & - & \scriptsize 96.6\% & \scriptsize 99.5\% (R$_1$RR) & \scriptsize 94.8\% (R$_1$RR) \\ \hline
\scriptsize Berretti \emph{et al.} (2013) \cite{Berretti:2013} & \scriptsize {3D} Face & \scriptsize 95.6\% & - & - & \scriptsize 97.3\% (R$_1$RR) & \scriptsize 92.8\% (R$_1$RR) \\ \hline
\scriptsize Berretti \emph{et al.} (2010) \cite{Berretti:2010} & \scriptsize {3D} Face & \scriptsize 94.15\% & - & - & \scriptsize $\approx$97.3\% (R$_1$RR) & \scriptsize $\approx$91.0\% (R$_1$RR) \\ \hline
\scriptsize Mohammadzade \emph{et al.} (2013) \cite{Mohammadzade:2013} & \scriptsize {3D} Face & - & - & \scriptsize {{99.2\%}} & - & - \\ \hline
\scriptsize Mian \emph{et al.} (2008) \cite{Mian:2008} & \scriptsize {3D} Face & \scriptsize 93.5\% & - & \scriptsize Neutral 99.9\% \newline Non-neutral 92.7\% & \scriptsize 99\% & \scriptsize 86.7\% \\ \hline
\scriptsize Mian \emph{et al.} (2007) \cite{Mian:2007} & \scriptsize {2D+3D} Face \newline {2D+3D} Nose & \scriptsize 95.91\% \newline $\approx$92.2\% & - & \scriptsize 99.3\% \newline 92.5\% & \scriptsize 99.2\%  \newline $\approx$94.9\% & \scriptsize 95.37\%  \newline $\approx$80.0\% \\ \hline
    \end{tabular}
\end{center}
\caption{Performance comparison on the FRGC dataset.}
\label{Tab:FRGCCompRes}
\end{table*}
\renewcommand{\baselinestretch}{2.0}

The results of applying different patches and curves to this dataset are given in Table \ref{Tab:FRGCCompRes}.
The table also compares the performance of the proposed approach with recently proposed 3D face recognition techniques that employ the nasal region and also the whole face. For the FRGC v2.0 experiment, the spherical patches outperform the curves, with a R$_1$RR of 97.9\%. For this experiment, although only the nasal region is used, the recognition rates of the proposed algorithm recognition rates are higher than or very close to the results of state-of-the-art approaches, in which the whole facial surface is utilised \cite{Smeets:2013,Osaimi:2009,Drira:2013,Alyuz:2010,Wang:20082,Li:2014,Berretti:2013,Mian:2007}. The recognition performance of recent methods that only use the nasal region, for example \cite{Wang:20082,Alyuz:2010, Mian:2007,Spreeuwers:2011} is at least $3\%$ worse, and in some cases up to $6\%$ lower. Also, when the samples used in the probe set are changed from neutral to non-neutral, the proposed algorithm has the lowest variation in the R$_1$RR for both curves and spherical patches ($\approx 0.06\%$ for the spherical descriptors). It is interesting to note that there is even a slight increase in the recognition rates when non-neutral samples are used as probes (Table \ref{Tab:FRGCCompRes}, last column), which shows how robustly the algorithm has learned the facial expressions. In comparison with the algorithm of Li \emph{et al.} \cite{Li:2014}, which also uses the normal vectors as the basis for its feature space, there is $\approx 1.5\%$ improvement in the R$_1$RR and nearly a 4\% increase in R$_1$RR when the non-neutral samples of FRGC are used as the probes.

ROC III curves are the FRGC's cross-seasonal verification scenario and these are plotted in Fig. \ref{Fig:FRGCCMCROC}-b using a logarithmic scale. The EER and 0.1\% FAR for the proposed approach are 2.4\% and 93.5\%, respectively. Although these results are higher than some previous {3D} nose and face recognition methods (see the fifth column in Table \ref{Tab:FRGCCompRes}), they are outperformed by a number of algorithms that use the whole face, which show a higher robustness for the verification scenario.
For example, the algorithms of \cite{Osaimi:2009,Mian:2008,Drira:2013,Spreeuwers:2011} have lower identification performance than the proposed algorithm but a better verification performance. One conclusion from this may be that, for verification scenarios, the whole facial region might provide a higher confidence level when matched with a claimed identity.

\subsubsection{Robustness against different facial expressions}
\label{sec:RobustnessAgainst}
To more extensively evaluate the face recognition performance over different facial expressions the BU-3DFE and Bosphorus datasets are used, as they both contain samples with known expression types. For the BU-3DFE results the Bosphorus dataset is used for feature selection and vice versa. 

For the first set of experiments, BU-3DFE is used for feature selection and the Bosphorus dataset for test. The gallery consists of 105 neutral samples, one for each subject, and all the remaining non-neutral and neutral samples are used as probes. A comparison with the previous results is provided in Table \ref{Tab:BosComp}. While most comparison approaches have used the whole {3D} face, the results of the nasal spherical patches is highly competitive when applied over the same dataset.

The second experiment is based on using various expressions for the probe samples. The Bosphorus dataset expressions include neutral, anger, disgust, fear, happiness, sadness, and surprise. To investigate the algorithm's performance in recognising probe samples with unseen facial expressions, for each subject one expression type is selected for the gallery samples and another for the probe.
The results are shown in Table \ref{Tab:ExpTypes}, where each column relates to a facial expression and the number of samples used in the probe set is given. The average recognition rank for the 76 subjects with more than one neutral sample per subject is reported for the neutral vs. neutral experiment, as shown in the last column.

The spherical patches again outperform the nasal curves. The disgust expression deforms the noses more significantly than the other expressions and produces the lowest recognition ranks. On the other hand, the feature space is most invariant for the surprise expressions, as 100\% of the samples are correctly recognised for this probe set. The algorithm is compared with that of Li \emph{et al.} \cite{Li:2011}, in which similar evaluations are performed. The spherical patches show higher robustness, in particular for disgust ($\approx 12\%$), anger and fear ($\approx 6\%$), happy ($\approx 3\%$), sadness and surprise ($\approx 1.5\%$). The results reported in \cite{Li:2011}, however, are $\approx 1\%$ better when a neutral expression is used for the probe samples.

\renewcommand{\baselinestretch}{1.0}
\begin{table}[!t]
\begin{center}
\begin{tabular}{| L{2.5cm} | L{3.5cm} | L{0.9cm} |}
    \hline
	\bf Algorithm & \bf Modality and size & \bf R$_1$RR  \\ \hline \hline
{{Spherical patches}} \newline {{Curves}}  & {3D} Nose (105/2797) & {95.35\%} \newline 86.1\% \\ \hline
Li \emph{et al.} (2014) \cite{Li:2014} & {3D} Face (105/2797) & 95.4\% \\ \hline
Dibeklio\u{g}lu \cite{Dibeklioglu:2009} & {3D} Nose (47/1527) \newline \ \ \ \ \ \ \ \ \ \ \ \ (47/423) [rotated] & 89.2\% \newline 62.6\% \\ \hline
Li \emph{et al.} (2011) \cite{Li:2011} & {3D} Face (105/4561)  & 94.1\% \\ \hline
Aly\"uz \emph{et al.} (2008) \cite{Alyuz:2008} & {3D} Face (34/441) \newline  \ \ \ \ \ \ \ \ \ \ \ (47/1508)  & 95.9\% \newline 95.3\% \\ \hline
    \end{tabular}
\end{center}
\caption{Comparison of some of the previous works on the 105 subjects in the Bosphorus dataset: the numbers in parentheses show the No. gallery samples/ No. of probe samples.}
\label{Tab:BosComp}
\end{table}
\renewcommand{\baselinestretch}{2.0}

\renewcommand{\baselinestretch}{1.0}
\begin{table*}[!t]
\begin{center}
\begin{tabular}{| L{2.5cm} | C{1.6cm} | C{1.7cm} | C{1.5cm} | C{1.7cm} | C{1.5cm} | C{1.6cm} | C{1.7cm} |}
\hline
\multirow{2}{*}{\bf Algorithm} & \multicolumn{7}{c|}{\bf Facial expression (No. of probe samples)} \\ \cline{2-8}
	 & {\bf{Happy}} \bf (106)  & {\bf{Surprise}} \bf (71) & {\bf{Fear}} \bf (70) & {\bf{Sadness}} \bf (66) & {\bf{Anger}} \bf (71) & {\bf{Disgust}} \bf (69) & {\bf{Neutral}} \bf (194) \\ \hline \hline
{{Spherical patches}} \newline {{Curves}} & 98.08\% \newline 85.85\% & 100\% \newline 92.96\% & 98.55\% \newline 87.14\% & 96.92\% \newline 92.31\% & 94.12\% \newline 84.06\% & 88.24\% \newline 69.12\% & 98.96\% \newline 96.88\%\\ \hline
Li \emph{et al.} \cite{Li:2011} & 95.28\% & 98.59\% & 92.86\% & 95.45\% & 88.73\% & 76.81\% & 100\% \\ \hline
    \end{tabular}
\end{center}
\caption{R$_1$RR for different expression types from the Bosphorus dataset used as probes, with 105 neutral samples used as the gallery.}
\label{Tab:ExpTypes}
\end{table*}
\renewcommand{\baselinestretch}{2.0}

\renewcommand{\baselinestretch}{1.0}
\begin{table*}[!t]
\begin{center}
\begin{tabular}{| L{2.5cm} | C{1.6cm} | C{1.8cm} | C{1.7cm} | C{1.8cm} | C{1.6cm} | C{1.7cm} |}
\hline
\multirow{2}{*}{\bf Algorithm} & \multicolumn{6}{c|}{\bf Facial expression (No. of probe samples)} \\ \cline{2-7}
	 & {\bf{Happy}} \bf (400)  & {\bf{Surprise}} \bf (400) & {\bf{Fear}} \bf (400) & {\bf{Sadness}} \bf (400) & {\bf{Anger}} \bf (400) & {\bf{Disgust}} \bf (400) \\ \hline \hline
{{Spherical patches}} \newline {{Curves}} & 88.5\% \newline 81.8\% & 91.0\% \newline 87.8\% & 89.8\% \newline 85.3\% & 92.3\% \newline 87.6\% & 90.1\% \newline 83.2\% & 81.8\% \newline 69.8\% \\ \hline
Hajati \emph{et al.} \cite{Hajati:2012} & 86.0\% & 84.0\% & 82.0\% & 85.0\% & 93.0\% & 79.0\% \\ \hline
    \end{tabular}
\end{center}
\caption{R$_1$RR for different expression types from the BU-3DFE dataset used as probes, with 100 neutral samples used as the gallery.}
\label{Tab:ExpTypesBU3DFE}
\end{table*}
\renewcommand{\baselinestretch}{2.0}

For a similar experiment with the BU-3DFE dataset, the 100 neutral samples are used as the gallery and the remaining
samples with different intensity levels of each expression used as probes. Table \ref{Tab:ExpTypesBU3DFE} shows the R$_1$RR for each expression type. As the expressions in the BU-3DFE dataset are significantly more intense than those in the Bosphorus dataset, the recognition rates are lower than those in Table \ref{Tab:ExpTypes}. 
The spherical patches still provide the best R$_1$RR performance for 5 of the 6 expression types, with an average R$_1$RR of
88.9\%, $\approx 4\%$ above that of \cite{Hajati:2012} which uses both the 2D and 3D information from the whole facial domain.


\section{Conclusions}
\label{sec:conc}
To address the problem of expression invariant face recognition, a novel algorithm is introduced, that utilises the {3D} shape of nose. The algorithm is based on a highly consistent and accurate landmarking algorithm, a robust feature space, discriminative feature descriptors and feature selectors. The proposed method is applied over three well-known face datasets, FRGC, BU-3DFE and Bosphorus. The matching results show that the algorithm is very successful for both the identification and verification scenarios, producing a R$_1$RR of 97.9\% on FRGC v2.0, an EER of 2.4\% on ROC III, and R$_1$RR of 98.45\% and 98.5\% for neutral and non-neutral probes, respectively. The proposed method does not rely on sophisticated preprocessing algorithms for its denoising and alignment. In addition, when there is only one sample per subject in the gallery, for all the merged folders of the FRGC dataset, a R$_1$RR of 96.2\% is obtained. For the Bosphorus dataset a R$_1$RR of 95.35\% is obtained when one neutral sample per subject is used for gallery and the remaining samples with various expression types as probes. The results of the proposed method reveal the high potential of the nasal region for {3D} face recognition. The recognition ranks are not only significantly higher than previous nasal region-based algorithms, but also have a better performance than many {3D} holistic and multi-modal approaches.

There are several aspects of the algorithm which can be utilised in other applications. For example, the feature extraction step, which is based on histograms of Gabor wavelet normals, can be applied to other {3D} object recognition methods. Also, the feature selection paradigm described here can be easily applied to other pattern recognition algorithms, to maximise the within-class and between-class similarity and dissimilarity, respectively, enabling the extraction of a lower dimensional and less redundant feature space. The application of the proposed landmarking algorithm can be investigated for performing facial alignment, low dimensional face recognition and pattern rejection. Finally, the application of the feature extraction step on the whole facial region, to make it robust against occlusions is an interesting area of future research.

\renewcommand{\baselinestretch}{1.0}
\bibliography{refs3}
\renewcommand{\baselinestretch}{2.0}





%
\renewcommand{\baselinestretch}{1.0}
\begin{IEEEbiography}[{\includegraphics[width=1in,height=1.25in,clip,keepaspectratio]{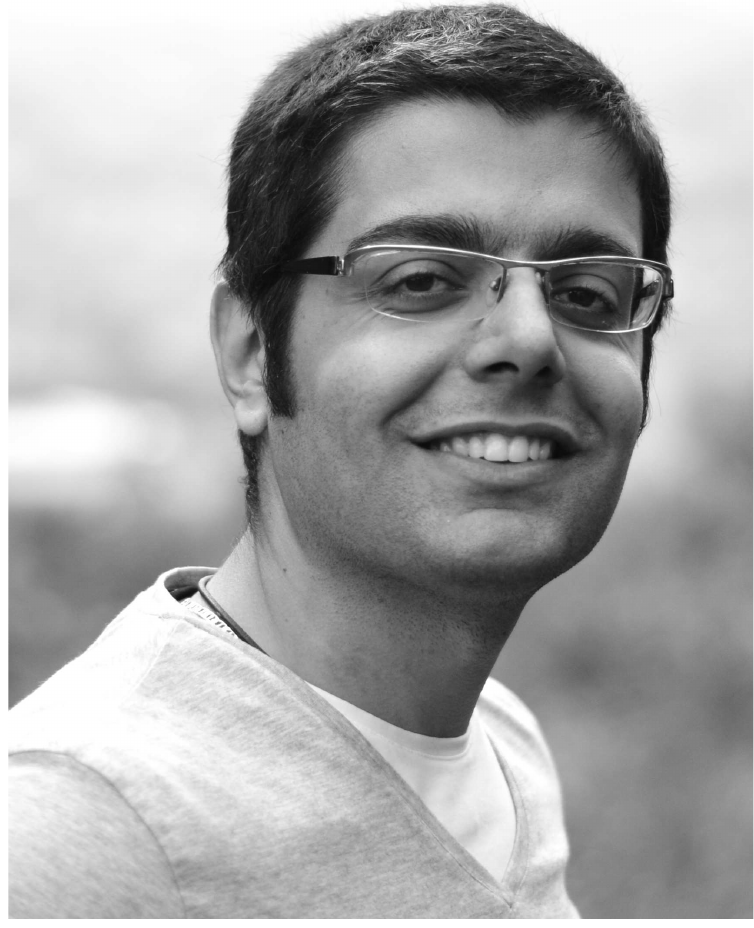}}]{Mehryar Emambakhsh}	
was awarded a PhD in Electronic and Electrical Engineering from the University of Bath in 2015, researching the potential of the {3D} shape of the nose for biometric authentication. He has been a Post-Doctoral Research Associate (PDRA) in big data science at Aston University and is currently a PDRA in RADAR and video data fusion for autonomous vehicles at Heriot-Watt University. His research interests are in {3D} object recognition and data mining.
\end{IEEEbiography}
\vspace{-1cm}
\begin{IEEEbiography}[{\includegraphics[width=1in,height=1.25in,clip,keepaspectratio]{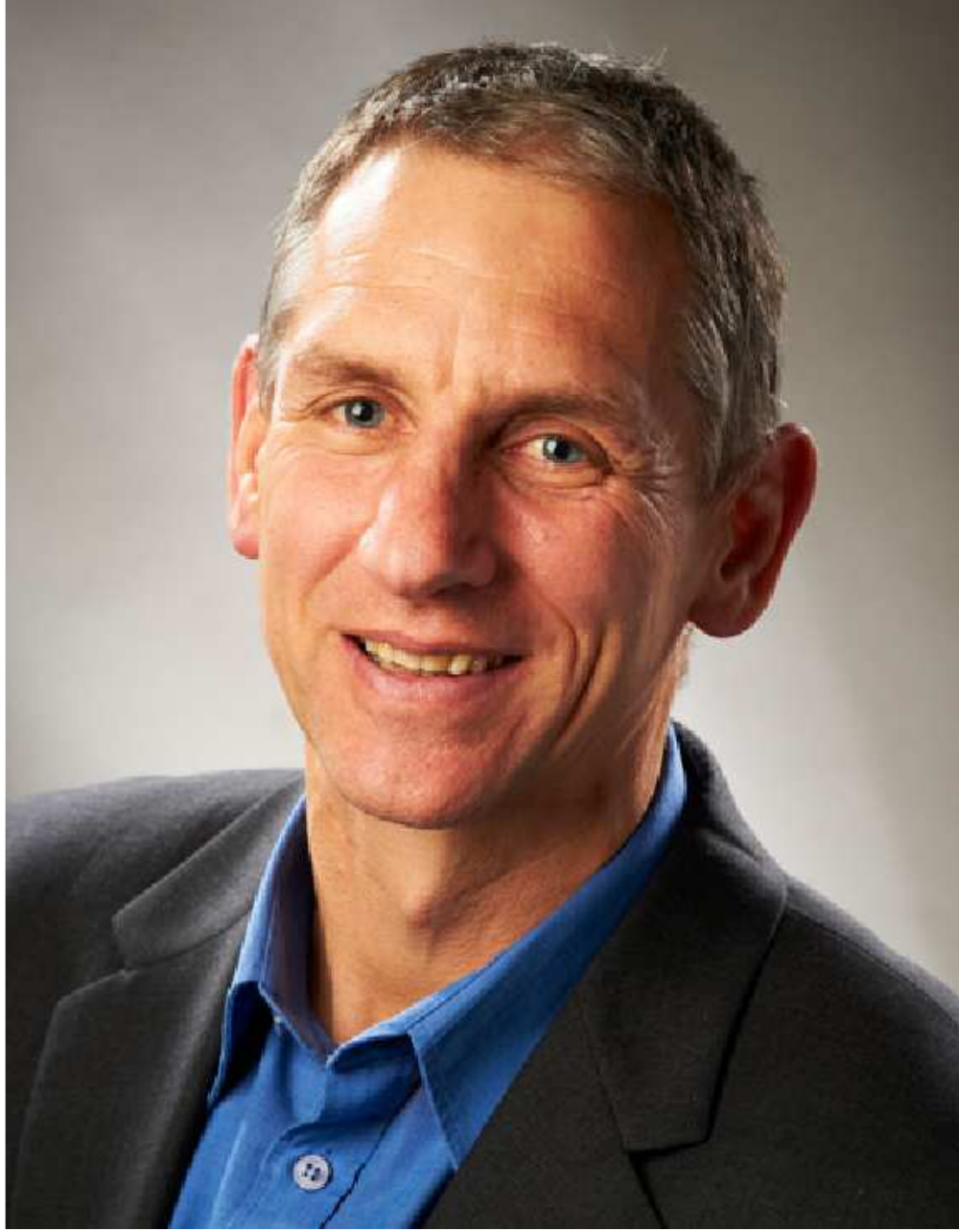}}]{Adrian Evans}
graduated from Loughborough University with a BEng in Electronics and Manufacturing Engineering in 1990 and was awarded a PhD in medical image processing from the University of Southampton in 1994. After a number of years as an academic at Massey University, New Zealand, he joined the Department of Electronic and Electrical Engineering at the University of Bath in 1997, where he is currently Head of Department. His research interests are in image and video processing and analysis and he has over 70 journal and refereed conference papers in this area. Current research projects include mathematical morphology, remote sensing, colour image processing and biometrics.
\end{IEEEbiography}
\renewcommand{\baselinestretch}{2.0}

\end{document}